%% file: IISE_Transactions_Latex_Final.tex
\documentclass[11pt]{article}
	
	\newcommand{\blind}{0}
	
    \makeatletter
    \renewcommand\section{\@startsection {section}{1}{\z@}%
                                       {-3.5ex \@plus -1ex \@minus -.2ex}%
                                       {2.3ex \@plus.2ex}%
                                       {\normalfont\fontfamily{phv}\fontsize{16}{19}\bfseries}}
    \renewcommand\subsection{\@startsection{subsection}{2}{\z@}%
                                         {-3.25ex\@plus -1ex \@minus -.2ex}%
                                         {1.5ex \@plus .2ex}%
                                         {\normalfont\fontfamily{phv}\fontsize{14}{17}\bfseries}}
    \renewcommand\subsubsection{\@startsection{subsubsection}{3}{\z@}%
                                        {-3.25ex\@plus -1ex \@minus -.2ex}%
                                         {1.5ex \@plus .2ex}%
                                         {\normalfont\normalsize\fontfamily{phv}\fontsize{14}{17}\selectfont}}
    \makeatother
	
	\usepackage{amsmath}
	\usepackage{graphicx}
	\usepackage{enumerate}
	\usepackage{natbib}
    \usepackage{subcaption}
	\usepackage{url} 
    \usepackage{amssymb}
    \usepackage{amsthm}
    
    \newtheorem{theorem}{Theorem}
    \newtheorem{remark}{Remark}
    \usepackage[caption=false,font=footnotesize]{subfig}
    
    \theoremstyle{definition}
    \newtheorem{assumption}{Assumption}
    \usepackage{algorithm}
    \usepackage{xcolor}
    \usepackage{booktabs}
    \usepackage{algpseudocode}
	
\begin{document}
		
		\def\spacingset#1{\renewcommand{\baselinestretch}%
			{#1}\small\normalsize} \spacingset{1}
		
		\if0\blind
		{
			\title{\bf DAG-based Penalized Clustering}
			\author{Honglin Du $^a$ and Muxuan Liang $^b$ Xiang Zhong $^c$\\
			$^a$ Department of Industrial and Systems Engineering, University of Florida, \\ Gainesville, FL 32611, USA \\
                          $^b$ Department of Biostatistics, MD Anderson Cancer Center, \\ Houston, TX 77030, USA \\
             $c$ Department of Industrial and Systems Engineering, University of Florida, \\ Gainesville, FL 32611, USA}
			\date{}
			\maketitle
		} \fi
\title{\bf A Unified Framework for Structure-Aware Clustering and Heterogeneous Causal Graph Learning}
\date{}
\maketitle

	\begin{abstract}
In complex multivariate systems, interactions among variables are defined by dependency structures, often encoded as directed acyclic graphs ($\text{DAGs}$). However, dependency structures can vary across subjects, and ignoring this structural heterogeneity introduces bias and obscures subpopulation-specific dependencies. To address this, we propose Directed Acyclic Graph–based Dependency Clustering via Alternating Direction Method of Multipliers (DAG-DC-ADMM), a unified framework built upon Structural Equation Modeling (SEM) that jointly learns cluster assignments and cluster-specific dependency structures. We encode acyclicity via a smooth constraint and integrate a groupwise truncated Lasso fusion penalty (gTLP) to cluster subjects based on their structural similarity. This yields a nonconvex optimization problem that incorporates sparsity, acyclicity, and structural consensus constraints. We address the nonconvexity by using the augmented Lagrangian method and solve it with an adapted version of the Alternating Direction Method of Multipliers (ADMM) for difference-of-convex programs. For certain graph structures, such as upper triangular adjacency matrices, our algorithm is guaranteed to converge to a Karush-Kuhn-Tucker (KKT) point. Experiments demonstrate that our method recovers cluster-specific causal dependency structures with a high true positive rate and a low false discovery rate. This capability enables the robust discovery of heterogeneous dependencies across subjects where the subpopulation label is unknown.

	\end{abstract}
			
	\noindent%
	{\it Keywords:} Directed Acyclic Graphs, Heterogeneous Causal Graphs, Structure-aware Clustering, Structural Equation Modeling.

	\spacingset{1.5} 

\section{Introduction} \label{s:intro}
In many multivariate systems where repeated measurements of individual variables are available, understanding how variables influence or depend on one another, i.e., dependency structures among variables, is crucial to uncovering the true mechanisms that drive the observed behavior. 
In this work, we define dependency structures as the patterns of conditional dependencies that characterize these influences. They are commonly encoded as directed acyclic graphs (DAGs) \citep{williams2018directed}. 
Properly estimating a DAG for a multivariate system often requires that samples be structurally homogeneous. However, this assumption can be violated, especially when data or measurements come from a mixture of unlabeled subpopulations or subjects. Estimating a single dependency structure from such data can introduce bias, leading to spurious connections or omission of true dependencies. In this work, we propose a unified framework that can automatically cluster subjects based on their underlying heterogeneous dependency structures and learn heterogeneous DAGs.

Towards heterogeneous causal graph learning, a naive solution is a two-step approach that first partitions subjects into homogeneous clusters and then learns a DAG separately in each cluster. However, a partition based on statistics such as means does not guarantee homogeneity in dependency structures. Consequently, traditional clustering methods, e.g., $K$-means \citep{Lloyd1982}, or hierarchical clustering \citep{murtagh2012algorithms,jain2010data} built upon these metrics may fail to distinguish distinct dependency patterns. An alternative two-step solution is to estimate the dependency structure for each subject, and then perform clustering based on these structures. Yet, in the settings where only a few measurements are collected per subject, the dependency structure estimated using limited data is susceptible to a lack of power. To overcome these limitations, we propose a unified approach that groups subjects by their dependency‐structure similarity to borrow information across subjects and subsequently gain efficiency of estimating heterogeneous dependency structures. 

Combining clustering with dependency‐structure learning is challenging. First, for problem formulation, in a unified framework, clustering and structure learning are not treated as separate or sequential tasks. We handle this by fusing an unsupervised clustering task with a supervised model-based structure learning task into a single structure-aware clustering problem in which cluster assignments are driven by similarity in the learned dependency structures. The dependency structures are parameterized using Structural Equation Models (SEMs) \citep{deng2018structural}, which express each feature as a linear function of its direct causes plus an independent disturbance term. This parametrization yields parameters that map transparently to causal effects. Crucially, the SEM parameterization allows us to define a closed-form and differentiable loss function, i.e., the reconstruction loss, 
which is combined with the groupwise truncated Lasso fusion penalty (gTLP), to form an objective function that encourages parameters to not only minimize loss but also form compact and well-separated clusters.
Furthermore, to ensure the learned structures are valid DAGs, we adopt the smooth acyclicity constraint proposed in NOTEARS (Non-combinatorial Optimization via Trace Exponential and Augmented Lagrangian for Structure Learning) \citep{zheng2018dags}, which enables gradient-based optimization over the space of DAGs, obviating a combinatorial search. We also impose a graph sparsity penalty to reduce spurious edges and enhance the interpretability of the learned graphs. This novel formulation simultaneously handles cluster assignments and causal-structure learning without sacrificing the performance of either task. 

Second, the resulting optimization problem is inherently nonconvex in the objective function and the domain, and demands an algorithm that guarantees solutions with provable convergence.
To solve this problem, we first address the nonconvexity term introduced by gTLP in the objective function. Following the difference-of-convex function algorithm \citep{wu2016new}, we express the objective function as the difference of two convex functions. Then, we construct a convex surrogate subproblem by replacing the subtracted convex function with its first-order affine minorization. Next, we incorporate the acyclicity constraints into the objective function via the augmented Lagrangian method, which employs Lagrange multipliers and quadratic penalties. Finally, we solve the resulting unconstrained difference-of-convex program with the Alternating Direction Method of Multipliers (ADMM) method. We refer to this as Directed Acyclic Graph–based Dependency Clustering via Alternating Direction Method of Multipliers (DAG-DC-ADMM), a unified structure-aware clustering and causal graph learning framework.

Under standard regularity conditions, we show that DAG-DC-ADMM converges to a stationary point of the augmented Lagrangian problem of the original constrained nonconvex optimization. Furthermore, when the adjacency matrices are upper-triangular,  our method is guaranteed to converge to a Karush-Kuhn-Tucker ($\text{KKT}$) point of the corresponding nonconvex optimization problem. Numerical studies show that DAG-DC-ADMM consistently achieves superior performance in structure recovery over population-level and individual-subject-based learning across various problem settings. In the case study, by jointly recovering cluster assignments and cluster-specific causal structures of human cell responses to biochemical perturbations, our method successfully identifies structural heterogeneity that is masked by methods based on homogeneity assumption. In summary, our contributions include 1) a structure-aware clustering and heterogeneous causal graph learning framework capable of capturing causal heterogeneity in complex systems, and 2) an effective algorithm to solve the proposed nonconvex optimization problem, together, offering a basis for dependency-driven clustering across diverse applications.

\textcolor{black}{In summary, this paper makes three main contributions. First, we introduce a structure-aware clustering perspective in which subjects are grouped according to similarities in their causal dependency structures rather than distance-based similarity in observed features or summary statistics. Second, we develop a unified DAG-constrained optimization framework that jointly estimates subject-specific dependency structures and induces clusters through a groupwise truncated Lasso fusion penalty, thereby avoiding the potential disconnect between clustering and structure learning in two-stage procedures. Third, we design a DC-ADMM algorithm for the resulting nonconvex problem and establish its convergence properties under appropriate regularity conditions. Numerical experiments and a real-data case study further show that the proposed framework can recover heterogeneous dependency patterns and produce interpretable subgroup-specific causal structures.}

The remainder of this paper is organized as follows. Section \ref{s:sec2} reviews relevant causal dependency structure learning and clustering methods. Section~\ref{s:methodology} introduces our proposed framework, which jointly estimates cluster assignments and dependency structures using the ADMM optimization strategy. Section~\ref{s:experiment} presents numerical experiments that evaluate the performance of our method under various scenarios. Section~\ref{s:case_study} provides a real-world case study to demonstrate the practical utility of our approach. Section ~\ref{s:discussion} concludes the paper and outlines potential directions for future research.

\section{Literature Review} \label{s:sec2}
We review the current state and limitations of causal dependency structure learning and clustering based on dependency-structure heterogeneity to position our proposed DAG-DC-ADMM framework.

Causal structure learning aims to uncover dependency structures among variables \citep{pearl2009causality}. This field evolves through methods such as SEM, Granger causality in time series \citep{eichler2012causal}, Bayesian networks as probabilistic DAG representations \citep{Kitson2023}, and nonlinear approaches based on kernel functions and neural networks \citep{marinazzo2008kernel,ke2019learning}. Among these methods, the dependency structure is commonly represented as a DAG. However, imposing acyclicity on DAGs traditionally requires combinatorial search, which is an NP-hard problem \citep{chickering1996learning}. To address this challenge, \citet{zheng2018dags} reformulates the combinatorial acyclicity constraint as a smooth penalty, enabling scalable gradient-based optimization over the space of DAGs. This reformulation has been subsequently extended to different applications, such as nonlinear models to capture complex dependencies \citep{zheng2020learning}, dynamic causal structures in time series data \citep{pamfil2020dynotears}, and graph neural network-based formulations that accommodate mixed variable types \citep{yu2019dag}.

Despite the progress in continuous-optimization approaches to causal structure learning, these methods typically assume a single causal model and thus a fixed joint distribution for all observations. As a result, they often struggle to recover dependency structures in heterogeneous settings \citep{huang2020causal}. This motivates an interest in learning heterogeneous dependency structures. Approaches in this direction include the Joint Graphical Lasso \citep{danaher2014joint}, which jointly estimates multiple pre-specified Gaussian graphical models by encouraging shared sparsity or edge similarity across groups. \textcolor{black}{Beyond settings with known group structure, mixture-based approaches have been proposed to capture latent heterogeneity. Early work by \citet{thiesson1998learning} uses mixture models to learn mixtures of DAGs via the Expectation--Maximization (EM) algorithm, assigning each subject to the component DAG that best explains it. More recent advances further investigate causal discovery under mixtures of DAGs, including methods for identifying causal relations from heterogeneous or nonstationary data \citep{strobl2023causal} and theoretical analysis of identifiability and separability conditions in mixture settings \citep{varici2024separability}. In parallel, other works focus on modeling subject-level heterogeneity in causal structures. For example, \citet{li2018learning} propose mixed-effects structural equation models to learn subject-specific DAGs by incorporating both shared and individual-specific components. From a Bayesian perspective, \citet{castelletti2023bayesian} develop nonparametric mixture models of Gaussian DAGs to capture heterogeneous causal effects across latent subpopulations.} In psychometrics, latent-class SEMs cluster individuals based on differences in SEM parameters, thereby capturing subgroups with distinct dependency structures \citep{muthen2002beyond}. These methods are parametric but usually require pre-specifying the number of mixture components or relying on known group labels  \citep{drton2017structure,huang2020causal,glymour2019review}. This motivates an alternative optimization-based formulation that can connect heterogeneous DAG learning with data-driven clustering.

When the data are unlabeled, unsupervised learning approaches are typically used. Traditional clustering algorithms such as $K$-means and hierarchical clustering operate on geometric similarity \citep{jain2010data,murtagh2012algorithms}. These methods do not rely on any parametric model assumptions for the data-generating process, and instead define clusters implicitly through pairwise distances between data points, which is typically less efficient. This motivates a shift towards fusion-penalized parametric clustering formulations, beginning with the sum-of-norms (SON) approach \citep{pelckmans2005convex,hocking2011clusterpath}. SON reformulates clustering by assigning each subject its own centroid and penalizing pairwise centroid differences with an $\ell_2$ fusion term. The fusion parameter controls how many centroids merge, mitigating the need to pre-specify the number of clusters. This formulation yields a convex optimization problem with a unique global optimal solution. However, convex fusion penalties often over-shrink and introduce bias. This leads to penalized-regression clustering methods such as PRclust, which use nonconvex penalties like the grouped truncated Lasso to obtain sharper cluster boundaries to separate the clusters \citep{pan2013cluster}. \citet{chi2015splitting} improve scalability by separating fusion and fidelity terms via variable-splitting methods, including the Alternating Minimization Algorithm and ADMM. \citet{hallac2015network} further extend this approach to graph-structured data with distributed ADMM updates. Despite the advances, these methods remain distance-based or centroid-based and do not model dependency structures among variables. 
Our research fills this gap, developing a unified parametric, structure-aware clustering framework building on recent advances in causal structure learning. Our proposed approach connects DAG-based causal discovery with unsupervised clustering and provides efficient solution approaches, charting a new direction for dependency-driven clustering.

\section{Methodology} \label{s:methodology}
\subsection{Problem Formulation}\label{s:methods.1}
In this study, we consider general scenarios where each subject may have multiple measurements, such as longitudinal data generated from a time-homogeneous stochastic process or multiple independently drawn samples from the same experiment. Multiple measurements, such as longitudinal data or repeated measurements, are common in medical and engineering applications. 
Let $\{\mathbf X_i\}_{i=1}^n$ denote the observed data from $n$ subjects, where each subject $\mathbf X_i$ contains $m_i$ independent and identically distributed (i.i.d.) measurements in a $d$-dimensional space. The data matrix for subject $i$ is
\begin{equation}
 \mathbf{X}_i
= \begin{bmatrix}
x_i^{(1,1)} & x_i^{(1,2)} & \dots & x_i^{(1,d)} \\
x_i^{(2,1)} & x_i^{(2,2)} & \dots & x_i^{(2,d)} \\
\vdots & \vdots & \ddots & \vdots \\
x_i^{(m_i,1)} & x_i^{(m_i,2)} & \dots & x_i^{(m_i,d)}
\end{bmatrix} \in \mathbb{R}^{m_i \times d}.
\end{equation}

Each $\mathbf X_i$ is modeled by a structural equation model (SEM):
\begin{equation}
\mathbf{X}_i = \mathbf{X}_i \mathbf{W}_i + \mathbf{Z}_i,
\qquad
\mathbf{X}_i,\,\mathbf{Z}_i \in \mathbb{R}^{m_i \times d},
\label{eq:SEM}
\end{equation}
where the adjacency matrix $\mathbf{W}_i \in \mathbb{R}^{d \times d}$ encodes the dependency structure among variables, and $\mathbf{Z}_i$ collects the exogenous disturbance terms. $\mathbf{Z}_i$ is assumed to have zero mean, i.e., $\mathbb{E}[\mathbf{Z}_i] = \mathbf{0}$ \citep{bollen2013eight}. \textcolor{black}{In this work, we do not impose a specific parametric distribution on $\mathbf{Z}_i$. Instead, we assume that the disturbance terms arise from the same distributional family across subjects. This assumption reflects our modeling focus that subject-level heterogeneity is driven primarily by differences in the structural matrices $\mathbf{W}_i$, rather than from differences in the  exogenous disturbance terms.} Our goal is to cluster subjects based on their underlying dependency structures while simultaneously recovering a set of distinct causal adjacency matrices $\mathbf{W}_i$. We define two subjects as belonging to the same cluster if their causal adjacency matrices $\mathbf{W}_i$ and $\mathbf{W}_j$ are identical. Let $\mathbf{W} = \{\mathbf{W}_i\}_{i=1}^n$ denote the collection of adjacency matrices. The objective function of the structural learning problem is then formulated as: 
\begin{equation}\min_{\;\mathbf{W}}\sum_{i=1}^n \frac{1}{2m_i}\| \mathbf{X}_i - \mathbf{X}_i \mathbf{W}_i \|_F^2 
  + \lambda_1 \sum_{i=1}^n \| \mathbf{W}_i \|_1.
  \label{eq:penalized_loss_function_0}
\end{equation}
The first term measures the reconstruction loss, and the second term encourages sparsity in each $\mathbf{W}_i$. The nonnegative parameter $\lambda_1$ serves as the regularization weight that controls the sparsity of each adjacency matrix $\mathbf{W}_i$. In addition to the reconstruction loss, other differentiable loss functions are also compatible with our framework, such as the negative log-likelihood for additive SEMs with Gaussian errors \citep{buhlmann2014cam}, or the logistic loss for generalized linear models \citep{zheng2020learning}.

The formulation \eqref{eq:penalized_loss_function_0} entails estimating subject-specific $\mathbf{W}_i$ individually, which may lead to a loss in efficiency if some subjects actually share common adjacency matrices or if $m_i$ is relatively small. To improve estimation efficiency and better identify clusters that share similar dependency structures, we estimate $\mathbf{W}$ by minimizing a penalized loss function that includes a fusion term:
\begin{equation}
\min_{\;\mathbf{W}}\sum_{i=1}^n \frac{1}{2m_i}\| \mathbf{X}_i - \mathbf{X}_i \mathbf{W}_i \|_F^2 
  + \lambda_1 \sum_{i=1}^n \| \mathbf{W}_i \|_1 
  + \lambda_2 P(\mathbf{W}).
\label{eq:penalized_loss_function}
\end{equation}
The term $P(\mathbf{W})$ is the fusion penalty that promotes clustering by penalizing pairwise differences among the $\{\mathbf{W}_i\}_{i=1}^n$. Assigning each subject its own $\mathbf{W}_i$ minimizes the reconstruction loss but can lead to poor generalization and interpretability. Conversely, forcing all $\mathbf{W}_i$ to be identical ignores structural heterogeneity and risks underfitting. The parameter $\lambda_2$ governs this balance: a larger $\lambda_2$ promotes fewer clusters, while smaller values allow more variation across clusters.

To avoid over-shrinking distinct dependencies, we use a nonconvex gTLP for $P(\mathbf{W})$: 
\[
P(\mathbf{W})=\sum_{i<j}\text{TLP}\bigl(\|\mathbf W_i-\mathbf W_j\|_F;\tau\bigr),
\]
where the $\text{TLP}$ is given by $\text{TLP}(\beta; \tau) = \min(|\beta|, \tau)$ for a scalar $\beta$ and a threshold parameter $\tau > 0$ \citep{wu2016new}. This threshold $\tau$ controls the extent of the fusion penalty: differences smaller than $\tau$ are penalized proportionally as an $\ell_1$-penalty, encouraging the fusion of similar $\mathbf{W}_i$. In contrast, larger differences are capped at $\tau$, allowing dissimilar matrices to remain distinct. We introduce auxiliary variables $\boldsymbol{\theta}_{ij}=\mathbf{W}_i-\mathbf{W}_j$, $1 \leq i < j \leq n$, which represent pairwise differences between adjacency matrices, and are referred to as the consensus constraints. \textcolor{black}{The auxiliary variables $\boldsymbol{\theta}_{ij}$ separate the representation of pairwise structural differences from the subject-specific adjacency matrices. This reformulation makes the subsequent optimization easier to organize by treating structural estimation and pairwise fusion as separate components.}

We also enforce acyclicity on the adjacency matrix of  each subject to rule out feedback loops and retain a clear causal interpretation. Specifically, we require each $\mathbf W_i$ to be a DAG by imposing the smooth constraint:
\[
h(\mathbf W_i)\;=\;\operatorname{tr}\!\big(\exp(\mathbf W_i\!\circ\!\mathbf W_i)\big)\;-\;d\;=\;0,
\quad i=1,\dots, n,
\]
where $\circ$ denotes the Hadamard product and $\operatorname{tr}(\cdot)$ the matrix trace.
This equality holds if and only if $\mathbf W_i$ is acyclic. Bringing together the reconstruction, sparsity, and clustering penalties, we arrive at the following constrained optimization problem over the adjacency matrices $\{\mathbf W_i\}_{i=1}^{n}$ and their pairwise differences
$\{\boldsymbol\theta_{ij}\}_{1\le i<j\le n}$:
\begin{equation}
\begin{split}
\min_{\{\mathbf{W}, \boldsymbol{\theta}\}} \mathcal{F}(\mathbf{W}, \boldsymbol{\theta}) = 
& \sum_{i=1}^n \frac{1}{2m_i}\| \mathbf{X}_i - \mathbf{X}_i \mathbf{W}_i \|_F^2 
+ \lambda_1 \sum_{i=1}^n \| \mathbf{W}_i \|_1 + \lambda_2 \sum_{i<j} \text{TLP}(\| \boldsymbol{\theta}_{ij} \|_F; \tau)\\
\text{s.t. }\;& h(\mathbf W_i)=0,\quad i=1,\dots, n,\\
& \boldsymbol\theta_{ij}=\mathbf W_i-\mathbf W_j,\quad 1\le i<j\le n.
\end{split}
\label{eq:rewrite_object_func_gTLP_augm}
\end{equation}

In summary, we model each subject using a linear SEM with a subject-specific DAG encoded by $\mathbf W_i$. We learn these causal adjacency matrices by minimizing a reconstruction loss, subject to sparsity and fusion penalties, and simultaneously enforcing acyclicity using the smooth constraint $h(\mathbf{W}_i)=0$. The gTLP fusion term clusters subjects by shrinking similar $\mathbf W_i$ together, yielding a structure-aware clustering of subjects according to their dependency structures.

\subsection{DAG-DC-ADMM Algorithm}\label{s:methods.2}
To solve the nonconvex problem in equation~\eqref{eq:rewrite_object_func_gTLP_augm}, we develop DAG-DC-ADMM, which extends the solution to penalized clustering by \citet{wu2016new}, tailored to two sources of nonconvexity: (i) the gTLP term in the objective and (ii) the acyclicity constraint on $\{\mathbf W_i\}_{i=1}^n$. We employ an iterative approach where difference-of-convex programming is used to handle the gTLP nonconvexity by linearizing the objective function's subtracted convex part at each iteration. Then, the resulting optimization problem, which remains nonconvex due to the acyclicity constraint, is solved by ADMM and the Augmented Lagrangian method.

\textcolor{black}{The gTLP penalty has a capped-linear form, which is naturally compatible with a DC representation: it can be written as a convex fusion term minus a convex correction term that prevents excessive shrinkage for sufficiently separated pairs. We therefore use difference-of-convex (DC) programming \citep{pan2013cluster,9325001} to split}
For source (i), \textcolor{black}{we use difference of convex
(DC) programming \citep{pan2013cluster,9325001} to split} the objective function $\mathcal F(\mathbf{W},\boldsymbol{\theta})$into a difference of two convex functions:
\[\mathcal{F}(\mathbf{W},\boldsymbol{\theta}) = \mathcal{F}_1(\mathbf{W},\boldsymbol{\theta})-\mathcal{F}_2(\boldsymbol{\theta}),\]
where $\mathcal{F}_1$ is fully convex, and $\mathcal{F}_2$ captures exactly the nonconvex portion of gTLP. Specifically, the components are defined as:
\[
\mathcal{F}_1(\mathbf{W}, \boldsymbol{\theta}) = \sum_{i=1}^n \frac{1}{2m_i}\| \mathbf{X}_i - \mathbf{X}_i \mathbf{W}_i \|_F^2 
+ \lambda_1 \sum_{i=1}^n \| \mathbf{W}_i \|_1 
+ \lambda_2 \sum_{i<j} \| \boldsymbol{\theta}_{ij} \|_F \\
\]
and
\[
\mathcal{F}_2(\boldsymbol{\theta}) = \lambda_2 \sum_{i<j} (\| \boldsymbol{\theta}_{ij} \|_F - \tau)_+,
\]
where
 $(\beta)_+ = \max(\beta, 0)$. At each difference-of-convex iteration $s$, we replace the subtracted term $\mathcal{F}_2(\boldsymbol{\theta})$ by its first-order affine minorant around the current estimate $\hat{\boldsymbol{\theta}}^{(s)}$. This yields the linearized approximation $\bar{\mathcal{F}}_2^{(s)}(\boldsymbol{\theta})$:
\[
\bar{\mathcal{F}}_2^{(s)}(\boldsymbol{\theta}) = \mathcal{F}_2(\hat{\boldsymbol{\theta}}^{(s)}) + \lambda_2 \sum_{i<j} \left( \| \boldsymbol{\theta}_{ij} \|_F - \| \hat{\boldsymbol{\theta}}_{ij}^{(s)} \|_F \right) \mathbb{I}\left( \| \hat{\boldsymbol{\theta}}_{ij}^{(s)} \|_F \geq \tau \right),
\]
where $\mathbb{I}(\cdot)$ denotes the indicator function. Substituting this minorant into the original function $\mathcal F_1 - \mathcal F_2$ gives an upper convex approximating function $\bar{\mathcal{F}}^{(s+1)}$ at iteration $s+1$:
\begin{equation}
\begin{split}
\bar{\mathcal{F}}^{(s+1)}(\mathbf{W}, \boldsymbol{\theta}) = 
& \sum_{i=1}^n \frac{1}{2m_i}\| \mathbf{X}_i - \mathbf{X}_i \mathbf{W}_i \|_F^2 
+ \lambda_1 \sum_{i=1}^n \| \mathbf{W}_i \|_1 \\
& + \lambda_2 \sum_{i<j} \| \boldsymbol{\theta}_{ij} \|_F \, \mathbb{I}\left( \| \hat{\boldsymbol{\theta}}_{ij}^{(s)} \|_F < \tau \right)
+ \lambda_2 \tau \sum_{i<j} \mathbb{I}\left( \| \hat{\boldsymbol{\theta}}_{ij}^{(s)} \|_F \geq \tau \right).
\end{split}
\end{equation}

We address source (ii) by incorporating the smooth acyclicity constraint, together with the consensus constraints in equation~\eqref{eq:rewrite_object_func_gTLP_augm}, into an augmented Lagrangian. This enables gradient-based updates of $\mathbf W$ within ADMM while maintaining acyclicity.
\begin{equation}
\begin{split}
\mathcal{L}_{\text{aug}}(\mathbf{W}, \boldsymbol{\theta}) = 
& \sum_{i=1}^n \frac{1}{2m_i}\| \mathbf{X}_i - \mathbf{X}_i \mathbf{W}_i \|_F^2 
+ \lambda_1 \sum_{i=1}^n \| \mathbf{W}_i \|_1 + \sum_{i=1}^n \left[ \alpha_i h(\mathbf{W}_i) + \frac{\rho_1}{2} h(\mathbf{W}_i)^2 \right]\\
& + \lambda_2 \sum_{i<j} \| \boldsymbol{\theta}_{ij} \|_F \mathbb{I}\left( \| \hat{\boldsymbol{\theta}}_{ij}^{(s)} \|_F < \tau \right)
+ \lambda_2 \tau \sum_{i<j} \mathbb{I}\left( \| \hat{\boldsymbol{\theta}}_{ij}^{(s)} \|_F \geq \tau \right) \\
& + \sum_{i<j} \langle \mathbf{Y}_{ij}, \boldsymbol{\theta}_{ij} - (\mathbf{W}_i - \mathbf{W}_j) \rangle 
+ \frac{\rho_2}{2} \sum_{i<j} \| \boldsymbol{\theta}_{ij} - (\mathbf{W}_i - \mathbf{W}_j) \|_F^2,
\end{split}
\label{eq:lagrangian_full}
\end{equation}
where $\boldsymbol{\alpha}=(\alpha_1,\dots,\alpha_n)$ is the vector of Lagrange multipliers enforcing the acyclicity constraints $h(\mathbf{W}_i)=0$, and $\{\mathbf{Y}_{ij}\}_{i<j}$ are matrix-valued Lagrange multipliers for the consensus constraints $\boldsymbol{\theta}_{ij}=\mathbf{W}_i-\mathbf{W}_j$. The nonnegative scalars $\rho_1$ and $\rho_2$ serve as penalty parameters for the respective constraint violations. We write
$
\langle \mathbf{A}, \mathbf{B}\rangle = \mathrm{trace}\bigl(\mathbf{A}^\top \mathbf{B}\bigr)
$
to denote the Frobenius inner product between two matrices of the same dimension. Finally, defining the scaled dual variables $\mathbf{U}_{ij} = \mathbf{Y}_{ij}/\rho_2$, we reformulate ADMM updates accordingly. \textcolor{black}{This
reparameterization combines the linear dual term and the quadratic penalty term into a single quadratic form. The scaled form is equivalent to the unscaled dual update but yields simpler ADMM update expressions.}

\begin{equation*}
\begin{split}
\widehat{\mathbf{W}}_i^{(s,k+1)} &= \arg\min_{\mathbf{W}_i} \;\frac{1}{2m_i} \| \mathbf{X}_i - \mathbf{X}_i \mathbf{W}_i \|_F^2 
+ \lambda_1 \| \mathbf{W}_i \|_1 + \hat{\alpha}_i^{(s,k)}\, h(\mathbf{W}_i) + \frac{\rho_1}{2} h(\mathbf{W}_i)^2 \\[1mm]
& + \frac{\rho_2}{2} \sum_{j > i} \left\| \hat{\boldsymbol{\theta}}_{ij}^{(s,k)} - (\mathbf{W}_i - \widehat{\mathbf{W}}_j^{(s,k)}) + \hat{\mathbf{U}}_{ij}^{(s,k)} \right\|_F^2 \\
& + \frac{\rho_2}{2} \sum_{j < i} \left\| \hat{\boldsymbol{\theta}}_{ij}^{(s,k)} - (\mathbf{W}_i - \widehat{\mathbf{W}}_j^{(s,k+1)}) + \hat{\mathbf{U}}_{ij}^{(s,k)} \right\|_F^2,
\end{split}
\end{equation*}
\begin{equation*}
\hat{\boldsymbol{\theta}}_{ij}^{(s,k+1)} = 
\arg\min_{\boldsymbol{\theta}_{ij}} \begin{cases} 
\lambda_2 \tau + \frac{\rho_2}{2} \left\| \boldsymbol{\theta}_{ij} - (\widehat{\mathbf{W}}_i^{(s,k+1)} - \widehat{\mathbf{W}}_j^{(s,k+1)}) + \hat{\mathbf{U}}_{ij}^{(s,k)} \right\|_F^2, & \text{if } \left\| \hat{\boldsymbol{\theta}}_{ij}^{(s)} \right\|_F \geq \tau; \\[2mm]
\lambda_2 \left\| \boldsymbol{\theta}_{ij} \right\|_F + \frac{\rho_2}{2} \left\| \boldsymbol{\theta}_{ij} - (\widehat{\mathbf{W}}_i^{(s,k+1)} - \widehat{\mathbf{W}}_j^{(s,k+1)}) + \hat{\mathbf{U}}_{ij}^{(s,k)} \right\|_F^2, & \text{if } \left\| \hat{\boldsymbol{\theta}}_{ij}^{(s)} \right\|_F < \tau;
\end{cases}
\end{equation*}
\begin{equation*}
\hat{\alpha}_i^{(s,k+1)} = \hat{\alpha}_i^{(s,k)} + \rho_1\, h\left(\widehat{\mathbf{W}}_i^{(s,k+1)}\right), \quad 1 \leq i \leq n,
\end{equation*}
\begin{equation}
\hat{\mathbf{U}}_{ij}^{(s,k+1)} = \hat{\mathbf{U}}_{ij}^{(s,k)} + \hat{\boldsymbol{\theta}}_{ij}^{(s,k+1)} - \left(\widehat{\mathbf{W}}_i^{(s,k+1)} - \widehat{\mathbf{W}}_j^{(s,k+1)}\right), \quad 1 \leq i < j \leq n,
\label{DC_ADMM_result}
\end{equation}
where $k$ indicates the $k$-th iteration in ADMM. Using the scaled dual variables $\hat{\mathbf{U}}_{ij}^{(s,k)}$ and the current difference 
$\Delta_{ij} = \widehat{\mathbf{W}}_i^{(s,k+1)} - \widehat{\mathbf{W}}_j^{(s,k+1)} - \hat{\mathbf{U}}_{ij}^{(s,k)}$, we apply a block soft thresholding operator for the gTLP to update $\hat{\boldsymbol{\theta}}_{ij}^{(s,k+1)}$: 
\begin{equation}
\hat{\boldsymbol{\theta}}_{ij}^{(s,k+1)}
=
\begin{cases}
\Delta_{ij}, 
&\text{if }\|\hat{\boldsymbol{\theta}}_{ij}^{(s-1)}\|_F \ge \tau,\\[2mm]
\operatorname{ST}\bigl(\Delta_{ij};\,\lambda_2/\rho_2\bigr),
&\text{if }\|\hat{\boldsymbol{\theta}}_{ij}^{(s-1)}\|_F < \tau,
\end{cases}
\label{eq:softthresholding}
\end{equation}  
where $\operatorname{ST}(\boldsymbol{\theta};\gamma) = (\|\boldsymbol{\theta}\|_F - \gamma)_+ \cdot \boldsymbol{\theta}/\|\boldsymbol{\theta}\|_F$ 
is the block soft-thresholding operator \textcolor{black}{\citep{parikh2014proximal}}. Blocks with norm below the threshold $\lambda_2/\rho_2$ are driven exactly to zero, eliminating weak deviations, while larger blocks are shrunk proportionally. When $\|\hat{\boldsymbol{\theta}}_{ij}^{(s-1)}\|_F\ge\tau$, the penalty becomes constant, so no further shrinkage is applied and large deviations are retained. The variables $\widehat{\mathbf W}_i^{(s,k+1)}$ and $\widehat{\boldsymbol\theta}_{ij}^{(s,k+1)}$ are updated according to standard ADMM (\ref{DC_ADMM_result}) until the ADMM primal and dual residuals fall below prescribed tolerances. We summarize the procedure in Algorithm~\ref{alg:dc-admm}.

\begin{algorithm}[ht]
\caption{DAG--DC--ADMM for SEM-based clustering with acyclicity constraint}
\label{alg:dc-admm}
\begin{algorithmic}[1]
\Require Observations $\{\mathbf X_i\}_{i=1}^{n}$; tuning parameters $(\lambda_1,\lambda_2,\tau)$; initial penalties $(\rho_1,\rho_2)$; outer tolerance $\varepsilon_{\text{out}}>0$
\State \textbf{Initialize:} $s\gets 0$; initialize $\widehat{\mathbf W}_i^{(0)}$, $\widehat{\boldsymbol\theta}_{ij}^{(0)}$, duals $\widehat{\boldsymbol\alpha}^{(0)}=\mathbf 0$, $\widehat{\mathbf U}_{ij}^{(0)}=\mathbf 0$ for all $1\le i<j\le n$; compute $\bar{\mathcal F}^{(0)}$
\While{$s=0$ \textbf{ or } $\bigl|\bar{\mathcal F}^{(s)}-\bar{\mathcal F}^{(s-1)}\bigr|>\varepsilon_{\text{out}}$}
    \State $s\gets s+1$
    \State Using ADMM to obtain
           $\widehat{\mathbf W}_i^{(s)}$, $\widehat{\boldsymbol\theta}_{ij}^{(s)}$, and updated duals
           $\widehat{\boldsymbol\alpha}^{(s)}$, $\widehat{\mathbf U}_{ij}^{(s)}$ in Equation~\eqref{DC_ADMM_result}.
    \State Update $\bar{\mathcal F}^{(s)}$
\EndWhile
\State \textbf{Output:} $\{\widehat{\mathbf W}_i\}_{i=1}^{n}$ and $\{\widehat{\boldsymbol\theta}_{ij}\}_{1\le i<j\le n}$; these estimates are then used to assign cluster labels.
\end{algorithmic}
\end{algorithm}
\begin{remark}
Good initial values are important to obtain meaningful shrinkage in the soft-thresholding step in (\ref{eq:softthresholding}). We initialize $(\widehat{\mathbf W}^{(0)}_i,\ \widehat{\boldsymbol\theta}^{(0)}_{ij})$ and the ADMM dual variables as follows. 
For each subject $i$, $\widehat{\mathbf W}_i^{(0)}$ is obtained by solving the $\ell_1$-regularized reconstruction problem
\[
\min_{\mathbf W_i}\ \frac{1}{2m_i}\bigl\|\mathbf X_i\mathbf W_i-\mathbf X_i\bigr\|_F^2
\;+\;\lambda_1\bigl\|\mathbf W_i\bigr\|_{1}
\quad\text{s.t.}\quad \operatorname{diag}(\mathbf W_i)=\mathbf 0,
\]
with the DAG and consensus penalties switched off (i.e., $\rho_1=\rho_2=0$). \textcolor{black}{In implementation, we solve this initialization problem using L-BFGS-B \citep{byrd1995limited}. We adopt a variable-splitting strategy \citep{zheng2018dags} and write $\mathbf W_i=\mathbf W_i^+-\mathbf W_i^-$ with $\mathbf W_i^+,\mathbf W_i^-\ge 0$, which converts the $\ell_1$ penalty into a smooth bound-constrained optimization problem.}
We then set the pairwise deviation variables as $
\widehat{\boldsymbol\theta}_{ij}^{(0)} \;=\; \widehat{\mathbf W}_i^{(0)}-\widehat{\mathbf W}_j^{(0)}$, $\widehat{\boldsymbol\theta}_{ji}^{(0)} \;=\; -\,\widehat{\boldsymbol\theta}_{ij}^{(0)},$
and initialize all ADMM dual variables to $\mathbf 0$.
This initialization is consistent with the consensus constraint and provides a well-scaled starting point. A poor choice of $\rho_1$ and $\rho_2$ can also slow down the ADMM subroutine and hence the overall DAG-DC-ADMM algorithm. We therefore adopt an adaptive strategy for the DAG-penalty parameter $\rho_1$. At each iteration, $\rho_1$ is increased by a constant factor $\eta_{\rho_1}>1$ \textcolor{black}{(e.g., $\eta_{\rho_1}=10$)} whenever the total acyclicity residual does not decrease sufficiently between iterations, namely, $h(\widehat{\mathbf W}^{(s,k+1)}) > \xi\, h(\widehat{\mathbf W}^{(s,k)}),$ where $\xi\in(0,1)$ is a tolerance parameter \textcolor{black}{(e.g., $\xi=0.25$). The parameter $\eta_{\rho_1}$ controls the rate at which the acyclicity penalty is increased, while $\xi$ determines when the current decrease in acyclicity violation is considered insufficient.} This schedule promotes faster early progress and stricter acyclicity enforcement.
In contrast, we keep the consensus-penalty parameter $\rho_2$ fixed, as moderate values of $\rho_2$ are empirically robust across datasets.
\end{remark}
After running Algorithm~\ref{alg:dc-admm}, each subject $i$ yields an estimated adjacency matrix $\widehat{\mathbf W}_i$ and pairwise auxiliary variables
$\widehat{\boldsymbol\theta}_{ij}$ that capture between-subject differences. Although the $\widehat{\mathbf W}_i$ estimates are driven toward proximity by the fusion penalty's shrinkage effect, they are not necessarily identical. Consequently, $\widehat{\boldsymbol\theta}_{ij}$ explicitly quantifies the remaining between-subject differences.
We form a symmetric dissimilarity matrix $\boldsymbol\Theta\in\mathbb R^{n\times n}$ with
entries $\Theta_{ij}=\|\widehat{\boldsymbol\theta}_{ij}\|_F$ and $\Theta_{ii}=0$.
We then perform complete–linkage hierarchical clustering~\citep{murtagh2012hierarchical} on $\boldsymbol\Theta$. 

Let $\mathcal C_u$ and $\mathcal C_v$ be two distinct clusters. 
The inter–cluster distance between them is defined as the maximum pairwise dissimilarity between their members: $D(\mathcal C_u,\mathcal C_v)
=\max_{i\in \mathcal C_u,\; j\in \mathcal C_v}\,\Theta_{ij}$. Clusters are obtained by cutting the dendrogram at height $\tau$: starting from singletons,
we iteratively merge the two clusters with the smallest inter–cluster distance while
$D(\mathcal C_u,\mathcal C_v)\le \tau$, and stop once all inter–cluster distances exceed $\tau$.
Subsequently, the resulting partition $\{\mathcal C_r\}_{r=1}^{R}$ consists of maximal sets whose
complete–linkage diameter (the maximum pairwise dissimilarity within the set) does not exceed $\tau$.
For each cluster $\mathcal C_r$, we define the cluster consensus matrix as 
$\widehat{\mathbf W}_{\mathcal C_r}
= \bigl|\mathcal C_r\bigr|^{-1}
\sum_{i\in \mathcal C_r}\widehat{\mathbf W}_i,$
optionally followed by hard–thresholding at level $\delta>0$.
Each subject $i\in \mathcal C_r$ is then assigned the consensus matrix
$\widehat{\mathbf W}_{\mathcal C_r}$, yielding the cluster–specific dependency
structures $\{\widehat{\mathbf W}_{\mathcal C_r}\}_{r=1}^{R}$.

\subsection{Theoretical Results}

In this section, we first analyze the computational complexity of the proposed DAG-DC-ADMM algorithm to assess its scalability for large-scale data. We then establish the convergence properties of the method under both the general and the strictly upper-triangular constrained problem formulations.

For the computational complexity analysis, in each ADMM iteration, the primary computational costs stem from three sources: updating adjacency matrices, enforcing acyclicity, and performing the consensus updates.
\textcolor{black}{Let $M=\sum_{i=1}^n m_i$ denote the total number of observations across all subjects. For each subject $i$, the reconstruction term in the $\mathbf W_i$-update can be written as
\[
\|\mathbf X_i\mathbf W_i-\mathbf X_i\|_F^2
=
\operatorname{tr}\!\left(\mathbf W_i^\top \mathbf X_i^\top \mathbf X_i \mathbf W_i\right)
-2\operatorname{tr}\!\left(\mathbf X_i^\top \mathbf X_i \mathbf W_i\right)
+\operatorname{tr}\!\left(\mathbf X_i^\top \mathbf X_i\right).
\]
Since each matrix $\mathbf X_i^\top \mathbf X_i \in \mathbb{R}^{d\times d}$ can be precomputed once, the one-time preprocessing cost for constructing all subject-level Gram matrices is $\sum_{i=1}^n \mathcal O(m_i d^2)=\mathcal O(Md^2)$. After this preprocessing step, the iterative updates involving the reconstruction term depend on $d\times d$ matrix operations rather than directly on the individual sample sizes $m_i$.}
Second, enforcing acyclicity using the smooth penalty $h(\cdot)$ on the weight matrices $\{\widehat{\mathbf W}_i\}_{i=1}^n$ requires a total cost of $\mathcal{O}(n d^3)$, which stems from calculating the matrix exponential and its gradient across all $n$ subjects. Finally, performing the consensus update requires operations over all $n(n-1)$ pairs of adjacency matrices. The current implementation incurs a cost of $\mathcal{O}(n^2 d^2)$ per iteration for this step. Therefore, one ADMM iteration costs \textcolor{black}{$\mathcal{O}\!\big(nd^3 + n^2 d^2\big)$}. With $K$ inner ADMM iterations nested inside $S$ difference-of-convex iterations, the total runtime is 
\textcolor{black}{$O(n m d^2 )+ \mathcal{O}\!\big(S K \,(nd^3 + n^2 d^2)\big)$}. \textcolor{black}{This overall complexity shows that the sequence length $m$ affects the algorithm only through a one-time preprocessing step, while the iterative computational burden is governed primarily by the number of subjects \(n\) and the dimension \(d\). Specifically, the computational cost is governed primarily by the acyclicity term \(\mathcal O(nd^3)\) and the quadratic consensus term \(\mathcal O(n^2d^2)\). The total cost is dominated by the acyclicity term when $d \gg n$, while the cost is primarily limited by the quadratic consensus term $\mathcal{O}(n^2 d^2)$ when $n$ is large. Domain-guided feature selection and pre-clustering or subsampling can help control $d$ and $n$ in practice, thereby improving tractability in clinical or engineering applications.}

On the convergence property, at each difference-of-convex iteration $s$, the augmented Lagrangian in (\ref{eq:lagrangian_full}) is nonconvex because it includes the smooth acyclicity term $h(\cdot)$. By leveraging nonconvex ADMM theory together with the Kurdyka Łojasiewicz (KL) property \citep{attouch2010proximal} of our objective function, we establish the following convergence result under the standing assumptions on the penalty parameters.

\begin{theorem}
\label{thm:convergence}
Fix a difference-of-convex iteration $s$ and consider $\mathcal{L}_{\mathrm{aug}}^{(s)}(\mathbf{W}, \boldsymbol{\theta})$ defined in~\eqref{eq:lagrangian_full}. Assume that $\mathcal{L}_{\mathrm{aug}}^{(s)}$ is a KL function and that the penalty parameters satisfy the standing assumptions (namely, $\rho_2>0$ is fixed and $\rho_1$ is nondecreasing and sufficiently large).
Let $\{(\mathbf W^{(s,k)},\boldsymbol\theta^{(s,k)})\}_{k\ge 0}$ be the sequence of ADMM iterates for this subproblem. Then the following conclusions hold. 

The sequence of iterates $\{(\mathbf W^{(s,k)},\boldsymbol\theta^{(s,k)})\}$ is guaranteed to be bounded. It has finite length, and the entire sequence converges to a first-order stationary point:
\[
(\mathbf W^{(s,k)},\boldsymbol\theta^{(s,k)}) \;\to\; (\mathbf W^{(s,\star)},\boldsymbol\theta^{(s,\star)}) \quad \text{as } k\to\infty.
\]
Here, $(\mathbf W^{(s,\star)},\boldsymbol\theta^{(s,\star)})$ denotes a stationary point of the ADMM subproblem for the fixed difference-of-convex iteration $s$.
\end{theorem}



While ADMM convergence to a stationary point of the augmented Lagrangian problem is guaranteed by Theorem \ref{thm:convergence}, the limit may not be a feasible solution to the original constrained problem and thus may not be a KKT point of the original constrained problem. In fact, for any feasible solution to the original constrained problem, unless it is also a stationary point of the unconstrained objective function, it cannot be a stationary point of the augmented Lagrangian objective. This result stems from a crucial property (Lemma 5 and Proposition 2 in \cite{wei2020dags}): for any feasible solution where the acyclicity constraint is exactly satisfied ($h(\mathbf{W})=0$), the constraint gradient vanishes, i.e., $\nabla_{\mathbf{W}} h(\mathbf{W}) = \mathbf{0}$. Thus, for any feasible solution, the KKT condition for the constrained problem is reduced to: 
$$\mathbf{0} \in \nabla_{\mathbf{W}}\left(  \sum_{i=1}^n \frac{1}{2m_i}\| \mathbf{X}_i - \mathbf{X}_i \mathbf{W}_i \|_F^2\right) + \partial_{\mathbf{W}} \left( \lambda_1 \sum_{i=1}^n \| \mathbf{W}_i \|_1 + \lambda_2 P(\mathbf{W}) \right) + \boldsymbol{\alpha} \nabla_{\mathbf{W}} h(\mathbf{W})$$
with $\nabla_{\mathbf{W}} h(\mathbf{W}) = \mathbf{0}$, which implies that the solution has to be a stationary point of the unconstrained objective function.
\begin{remark}
Despite lacking an optimality guarantee, the method is empirically effective. The NOTEARS paper demonstrates that the stationary points found by their method are comparable to the globally optimal solution (see Table 1 of \cite{zheng2018dags}). Given that our DAG-DC-ADMM framework leverages the same smooth acyclicity constraint,  the resulting solutions are also high-quality, as demonstrated in our numerical analysis.
\end{remark}

While Theorem \ref{thm:convergence} addresses the fully nonconvex formulation, the optimization problem simplifies significantly when a causal ordering is known a priori. Given this ordering, we restrict the adjacency matrix to be strictly upper triangular, thereby excluding cycles by construction, noting that the adjacency matrix of any DAG can be arranged into an upper triangular matrix by reordering the vertices according to a topological sort. When such ordering information is available, the explicit acyclicity penalty term $h(\mathbf{W})$ becomes redundant. This structural imposition simplifies the difference-of-convex subproblems, allowing us to leverage stronger convergence statements from convex optimization theory. 
\begin{assumption}[Strict upper-triangular structural prior]
\label{assump:upper}
Variables are indexed in a fixed order $1,\dots,d$, and $(\mathbf W_i)_{ab}$ denotes the weight on the edge $a\!\to\! b$.
Let
\[
\mathcal U
=\bigl\{\mathbf W\in\mathbb R^{d\times d}:\ \operatorname{tril}(\mathbf W,0)=\mathbf 0\bigr\}
=\bigl\{\mathbf W:\ W_{ab}=0\ \text{for all } a\ge b\bigr\}.
\]
\textcolor{black}{Here, $\operatorname{tril}(\mathbf W,0)$ denotes the lower triangular part of $\mathbf W$ including the main diagonal.} Impose $\mathbf W_i\in\mathcal U$ for all $i=1,\dots, n$.
Only entries with $a<b$ are permitted to be nonzero.
Under this constraint, every feasible $\mathbf W_i$ is acyclic; hence the acyclicity term $h(\mathbf W_i)$ is omitted.
\end{assumption}

Under Assumption~\ref{assump:upper}, the set
$\{\mathbf W_i:\operatorname{tril}(\mathbf W_i,0)=\mathbf 0\}$ is a convex linear subspace, and the constraints
$\{\boldsymbol{\theta}_{ij}=\mathbf W_i-\mathbf W_j\}_{1\le i<j\le n}$ are affine. Therefore, the feasible region is convex. The resulting optimization problem is
\begin{equation}
\label{eq:rewrite_object_func_gTLP_augm_upper}
\begin{split}
\min_{\mathbf W,\boldsymbol{\theta}}\ \mathcal F(\mathbf W,\boldsymbol{\theta})
=\;& \sum_{i=1}^{n}\frac{1}{2m_i}\bigl\|\mathbf X_i-\mathbf X_i\mathbf W_i\bigr\|_F^2
\;+\;\lambda_1\sum_{i=1}^{n}\|\mathbf W_i\|_{1}
\;+\;\lambda_2\sum_{i<j}\operatorname{\text{TLP}}\!\bigl(\|\boldsymbol{\theta}_{ij}\|_F;\tau\bigr)\\
\text{\rm s.t. }\;& \operatorname{tril}(\mathbf W_i,0)=\mathbf 0,\quad i=1,\dots, n,\\
& \boldsymbol{\theta}_{ij}=\mathbf W_i-\mathbf W_j,\quad 1\le i<j\le n.
\end{split}
\end{equation}
Within the difference-of-convex step, replacing $\operatorname{\text{TLP}}(\cdot;\tau)$ by its convex majorizer yields a convex approximation objective over this convex feasible region; consequently, the resulting subproblem can be solved as a convex program.

\begin{theorem}
\label{thm:finite}
Let $\{(\mathbf W^{(s)},\boldsymbol{\theta}^{(s)})\}_{s\ge 0}$ be the difference-of-convex outer-iteration sequence of DAG-DC-ADMM, and let $\mathcal F(\mathbf W,\boldsymbol{\theta})$ denote the difference-of-convex  objective under Assumption~\ref{assump:upper}.
Then there exists a finite index $s^\star < \infty$  such that
\[
\mathcal F\!\bigl(\mathbf W^{(s)},\boldsymbol{\theta}^{(s)}\bigr)
=
\mathcal F\!\bigl(\mathbf W^{(s^\star)},\boldsymbol{\theta}^{(s^\star)}\bigr),
\qquad s\ge s^\star,
\]
and $(\mathbf W^{(s^\star)},\boldsymbol{\theta}^{(s^\star)})$ is a KKT point.
\end{theorem}

This result highlights that incorporating domain knowledge into the causal structure not only simplifies the problem but also strengthens the theoretical optimality guarantees. The proofs of the theorems can be found in the Supplemental Material.

\section{Experiments}\label{s:experiment}
To assess the performance and robustness of the proposed DAG–DC–ADMM approach,  we perform a set of experiments,  tuning regularization parameters $\lambda_1$, $\lambda_2$ and $\tau$ using cross-validation. Specifically,  
$\lambda_{1}$ weights the $ \ell_1 $ sparsity penalty on the adjacency matrix of each subject. 
$\lambda_{2}$ weights the fusion penalty that encourages similar matrices to merge into the same cluster and $\tau$ defines the truncation threshold parameter that controls the shrinkage of pairwise differences between matrices. In the following subsections, we first introduce our experimental design, which consists of two data regimes (small-sample/long-series vs. large-sample/short-series), along with the grid of $\lambda_1$, $\lambda_2$, and $\tau$ values. Next, we describe how we generate synthetic data, detailing cluster assignments, DAG topologies, edge-weight distributions, and choices of the exogenous disturbance 
$\mathbf Z$ that represents unexplained variability. Then, we specify evaluation metrics in two aspects as clustering accuracy and structure-recovery accuracy. Finally, we present results across settings and threshold sweeps.

\subsection{Setup}
We examine six experimental settings defined by two sampling regimes—small-sample/long-series ($n = 50$ subjects, $m = 300$ measurements each) and large-sample/short-series ($n = 200$ subjects, $m = 50$ measurements each)—combined with three levels of structural disturbance variance $\sigma \in \{0.5, 1, 2\}$. 
In all cases, each subject has the same number of measurements $m_i = m$, and the disturbance terms are generated as $\mathbf{Z}_i \sim \mathcal{N}\bigl(\mathbf{0}, \sigma^2 \mathbf{I}_d\bigr)$. 
This design allows us to systematically evaluate how the proposed algorithm performs under varying cohort sizes, per-subject sequence lengths, and disturbance variances, and how these factors influence clustering performance and structure recovery.

For each scenario, we proceed as follows. 
\textcolor{black}{We consider $R_{\text{truth}} = 2$ clusters, where $R_{\text{truth}}$ denotes the true number of clusters} and partition the $n$ subjects into two groups according to proportions $[0.6,\,0.4]$. 
Let $\mathcal{C}_r$ denote the set of subjects in cluster $r$, for $r \in \{1,2\}$. 
For each cluster $\mathcal{C}_r$, we construct a cluster-specific DAG encoded by an adjacency matrix $\mathbf{W}_{\mathcal{C}_r} \in \mathbb{R}^{d \times d}$. 
To guarantee acyclicity, we select a pre-specified number of edges from the strict upper-triangular index set $\{(a,b) : a < b\}$, assign each selected edge a weight drawn uniformly from $[-0.5, -0.2] \cup [0.2, 0.5]$, and then apply a random node permutation to remove any implicit ordering. 
Next, for each subject $i \in \mathcal{C}_r$, we simulate an independent data matrix $\mathbf{X}_i \in \mathbb{R}^{m_i \times d}$ according to the SEM in \eqref{eq:SEM}:
\[
\mathbf{X}_i = \mathbf{Z}_i (\mathbf{I} - \mathbf{W}_{\mathcal{C}_r})^{-1}.
\]
Finally, we randomly permute the entire set of subjects, i.e., the matrices $\{\mathbf{X}_i\}_{i=1}^n$, 
their true cluster labels, and the corresponding generating DAGs $\{\mathbf{W}_{\mathcal{C}_r}\}_{r=1}^2$, to eliminate any residual ordering before applying the DAG–DC–ADMM algorithm.

We perform $50$ independent repetitions of the entire pipeline for each scenario. 
In each repetition, we first generate a new dataset $\{\mathbf{X}_i\}_{i=1}^n$ under the specified sampling regime and disturbance scale, and then conduct three-fold cross-validation over a predefined grid of $(\lambda_1, \lambda_2, \tau)$. 
Because our goal is to assign each subject to a cluster, we do not split subjects into disjoint train-test sets. 
Instead, we partition the $m_i$ rows of each $\mathbf{X}_i$ into three equally sized parts: in each fold, one part per subject forms the validation set $\{\mathbf{X}_{i,\mathrm{val}}\}_{i=1}^n$, and the remaining two parts form the training set. 
Within the cross-validation loop, each hyperparameter triplet yields candidate cluster-specific adjacency matrices $\{\widehat{\mathbf{W}}_{\mathcal{C}_r}\}_{r=1}^{\widehat{R}}$, \textcolor{black}{where $\widehat{R}$ denotes the estimated number of clusters.} Let $r(i)$ denote the assigned cluster of subject $i$'s current fold. We compute the validation reconstruction loss as
\[
\mathcal{L}_{\mathrm{rec}}
= \frac{1}{n} \sum_{i=1}^{n} \frac{1}{2m_i}
\bigl\| \mathbf{X}_{i,\mathrm{val}} - \mathbf{X}_{i,\mathrm{val}} \widehat{\mathbf{W}}_{\mathcal{C}_{r(i)}} \bigr\|_{F}^{2}.
\]

The hyperparameters achieving the lowest average validation loss are selected, after which we refit the DAG–DC–ADMM model on the full dataset for that repetition and record both clustering accuracy and structure-recovery accuracy. 
 Clustering quality is evaluated using the Adjusted Rand Index (ARI), homogeneity, and completeness. 
Given a reference partition and an algorithm-induced partition, ARI measures pairwise agreement between two clusterings, with an adjustment that removes the similarity expected from random label assignments. It equals $1$ for identical partitions and is approximately~$0$ under random assignment~\citep{hubert1985comparing}. 
Homogeneity measures cluster purity: whether each predicted cluster contains only members of a single reference group. Completeness measures class aggregation: whether all members of a reference group are assigned to the same predicted cluster~\citep{rosenberg2007v}. 
Both scores range from~$0$ to~$1$; high homogeneity with low completeness typically indicates over-segmentation, while the converse suggests over-merging. 
In our synthetic setting, the reference partition corresponds to the ground-truth labeling, and the induced partition is the labeling predicted by our method. Structure recovery is evaluated in two stages. 
In the skeleton stage, for each cluster-specific estimate $\widehat{\mathbf{W}}_{\mathcal{C}_r}$, we binarize entries and ignore directions to extract an undirected skeleton $\widehat{\mathbf{B}}_{\mathcal{C}_r}$. 
Given a sparsity threshold $\delta > 0$, for any distinct nodes $a \neq b$, we define
\[
(\widehat{\mathbf{B}}_{\mathcal{C}_r})_{ab}
= \mathbf{I}\!\left(
  \max\!\left\{
    \big|(\widehat{\mathbf{W}}_{\mathcal{C}_r})_{ab}\big|,
    \big|(\widehat{\mathbf{W}}_{\mathcal{C}_r})_{ba}\big|
  \right\} \ge \delta
\right),
\qquad
(\widehat{\mathbf{B}}_{\mathcal{C}_r})_{ba} = (\widehat{\mathbf{B}}_{\mathcal{C}_r})_{ab},
\qquad
\mathrm{diag}(\widehat{\mathbf{B}}_{\mathcal{C}_r}) = \mathbf{0}.
\]
That is, we place an undirected edge between nodes~$a$ and~$b$ if the magnitude of either directional weight is at least~$\delta$; otherwise, the edge is set to~$0$. Let the corresponding ground-truth DAG for cluster~$r$ be $\mathbf{W}_{\mathcal{C}_r}$ and its undirected skeleton $\mathbf{B}_{\mathcal{C}_r}$. 
For any subject~$i$ assigned to cluster~$r$, we evaluate 
(i)~the undirected skeleton $\widehat{\mathbf{B}}_{\mathcal{C}_r}$ against $\mathbf{B}_{\mathcal{C}_r}$ and 
(ii)~the directed matrix $\widehat{\mathbf{W}}_{\mathcal{C}_r}$ against $\mathbf{W}_{\mathcal{C}_r}$. 
From these comparisons, we compute the true positive rate (TPR), the false discovery rate (FDR), and the true negative rate (TNR). Here, TPR is the proportion of true edges correctly identified, FDR is the proportion of identified edges that are false, and TNR is the proportion of true non-edges correctly identified. 
The final reported scores are the macro-averages over all~$n$ subjects. Although high DAG recovery typically implies strong skeleton recovery, reporting skeleton metrics separately allows us to distinguish errors due to missing or spurious links from those due to incorrect edge orientations. 
Together, the skeleton and directed metrics quantify the method’s ability both to detect the presence of associations and to recover their correct causal directions.

Across all $50$ repetitions, we report the mean results for each scenario and compare them against the three NOTEARS-based benchmarks.
\begin{itemize}
  \item Population: learn a single DAG from all subjects' data pooled together.
  \item Individual: learn one DAG per subject by applying NOTEARS separately to each $\mathbf{X}_i$.
  \item Oracle: given the true cluster labels, aggregate data within each cluster and learn one DAG per cluster. 
\end{itemize}

These three NOTEARS-based methods represent distinct modeling scenarios. 
The population baseline corresponds to a homogeneous setting, where data from all subjects are pooled as if drawn from a single distribution. This ignores individual variability. 
The individual baseline represents the fully heterogeneous case, fitting one DAG per subject without information sharing and risking high variance and overfitting. 
The oracle cluster benchmark assumes perfect knowledge of the true clusters, pools data within each true group, and learns one DAG per cluster. Although infeasible in practice, it serves as an oracle upper bound on performance when the cluster label is known.  
We limit baselines to NOTEARS variants because our method adopts the same smooth acyclicity constraint, enabling a controlled comparison under shared optimization constraints and assumptions. Methods from other paradigms (e.g., the Peter Clark (PC) algorithm, Fast Causal Inference (FCI), and Greedy Equivalence Search (GES)) optimize different objectives and rely on distinct data assumptions, which would confound attribution of gains. Moreover, NOTEARS, as a continuous approach to DAG learning, has already been extensively benchmarked against these families and has shown competitive performance. Focusing our evaluation on it therefore isolates the effect of the proposed clustering and consensus components.

\subsection{Results}
For all comparable methods, the only hyperparameter is the sparsity parameter~$\lambda_1$, which is selected via three-fold cross-validation on held-out measurements within each subject.
To tune the proposed algorithm, we perform three-fold cross-validation over the following grid:
\[
\lambda_{1} \in \{10^{-4}, 10^{-3}, 10^{-2}, 10^{-1}\}, \quad
\lambda_{2} \in \{10^{-5}, 10^{-4}, 10^{-3}, 10^{-2}\}, \quad
\tau \in \{0.05, 0.1, 0.4, 0.7\}.
\]
We fix $\rho_{2} = 0.05$, while $\rho_{1}$ is initialized at $0.1$ and adaptively increased during optimization. 
The final hyperparameter combination is selected by minimizing the average reconstruction loss across validation folds.

\subsubsection{Small-sample/Long-series scenario}

In this scenario, we use $n = 50$ subjects, each with $m = 300$ independent measurements, and consider three disturbance scales $\sigma \in \{0.5, 1, 2\}$.

\begin{table}[ht]
\centering
\small
\caption{Clustering Results (Small-sample/Long-series, $n=50$, $m=300$). Values are mean $\pm$ 95\% CI over 50 runs; 
The true number of clusters $R_{\text{truth}}=2$ is fixed, and estimated $\widehat{R}$ is reported as the mode over repetitions.}
\setlength{\tabcolsep}{3pt} 
\begin{tabular}{lcccccccc}
\toprule
Setting & ARI & Homogeneity & Completeness & $\text{Recon.\ Error}_{\text{truth}}$& $\text{Recon.\ Error}_{\text{est}}$ & $R_{\text{truth}}$ & $\widehat{R}$ \\
\midrule
$\sigma=0.5$ & $0.91 \pm 0.03$ & $1.00 \pm 0.00$ & $0.85 \pm 0.04$ & $1.25 \pm 0.00$ & $1.25 \pm 0.00$ & 2 & 2 \\
$\sigma=1$   & $0.81 \pm 0.05$ & $1.00 \pm 0.00$ & $0.71 \pm 0.05$ & $5.00 \pm 0.01$ & $4.98 \pm 0.01$ & 2 & 3 \\
$\sigma=2$   & $0.83 \pm 0.05$ & $1.00 \pm 0.00$ & $0.76 \pm 0.05$ & $20.01 \pm 0.03$ & $19.90 \pm 0.03$ & 2 & 2 \\
\bottomrule
\end{tabular}
\label{tab:clust_selected_metrics_N502C}
\end{table}

Across 50 repetitions (Table~\ref{tab:clust_selected_metrics_N502C}), the method attains consistently high cluster purity (homogeneity $\approx 1.00$). Completeness dips at a higher disturbance level (e.g., $\sigma=1$), indicating mild over-segmentation in which each true cluster is split into smaller subpopulations. ARI remains high ($> 0.8$), reflecting stable clustering, and reconstruction errors increase monotonically with $\sigma$, as expected. The reconstruction errors computed under the true and estimated cluster assignments are nearly identical across all $\sigma$ levels. Table~\ref{tab:combined_recovery_metrics_N50} then reports structure-recovery metrics for the proposed DAG-DC-ADMM approach and the three NOTEARS‐based benchmarks. 

\begin{table}[htbp]
\centering
\small 
\caption{Skeleton and DAG recovery metrics for different methods under varying disturbance levels ($\sigma$). (Small‐sample/Long‐series, $n=50,\;m=300$; mean $\pm$ 95\% CI over 50 runs, threshold $=0.02$).}
\label{tab:combined_recovery_metrics_N50}
\setlength{\tabcolsep}{3pt} 
\begin{tabular}{l|c|r|r|r|r|r}
\toprule
Method & Skel TPR & Skel FDR & Skel TNR & DAG TPR & DAG FDR & DAG TNR \\
\midrule
\multicolumn{7}{l}{\textbf{Disturbance Scale $\sigma = 0.5$}} \\
\midrule
Population  & $0.94 \pm 0.03$ & $0.47 \pm 0.02$ & $0.17 \pm 0.05$ & $0.64 \pm 0.04$ & $0.64 \pm 0.02$ & $0.62 \pm 0.02$ \\
Individual  & $1.00 \pm 0.00$ & $0.21 \pm 0.00$ & $0.70 \pm 0.01$ & $0.81 \pm 0.02$ & $0.36 \pm 0.01$ & $0.84 \pm 0.01$ \\
DAG-DC-ADMM          & $1.00 \pm 0.00$ & $0.16 \pm 0.03$ & $0.77 \pm 0.04$ & $0.94 \pm 0.02$ & $0.30 \pm 0.03$ & $0.85 \pm 0.02$ \\
Oracle & $1.00 \pm 0.00$ & $0.08 \pm 0.02$ & $0.89 \pm 0.03$ & $0.88 \pm 0.03$ & $0.18 \pm 0.04$ & $0.92 \pm 0.02$ \\
\midrule
\multicolumn{7}{l}{\textbf{Disturbance Scale $\sigma = 1.0$}} \\
\midrule
Population  & $0.95 \pm 0.03$ & $0.46 \pm 0.02$ & $0.18 \pm 0.05$ & $0.64 \pm 0.04$ & $0.63 \pm 0.03$ & $0.63 \pm 0.02$ \\
Individual  & $1.00 \pm 0.00$ & $0.36 \pm 0.00$ & $0.40 \pm 0.01$ & $0.80 \pm 0.02$ & $0.49 \pm 0.01$ & $0.73 \pm 0.01$ \\
DAG-DC-ADMM          & $1.00 \pm 0.00$ & $0.14 \pm 0.02$ & $0.80 \pm 0.01$ & $0.93 \pm 0.02$ & $0.28 \pm 0.03$ & $0.86 \pm 0.02$ \\
Oracle & $1.00 \pm 0.00$ & $0.07 \pm 0.02$ & $0.91 \pm 0.03$ & $0.88 \pm 0.03$ & $0.17 \pm 0.04$ & $0.93 \pm 0.02$ \\
\midrule
\multicolumn{7}{l}{\textbf{Disturbance Scale $\sigma = 2.0$}} \\
\midrule
Population  & $0.96 \pm 0.03$ & $0.46 \pm 0.02$ & $0.19 \pm 0.04$ & $0.66 \pm 0.04$ & $0.62 \pm 0.03$ & $0.63 \pm 0.02$ \\
Individual  & $1.00 \pm 0.00$ & $0.29 \pm 0.01$ & $0.57 \pm 0.01$ & $0.80 \pm 0.02$ & $0.42 \pm 0.01$ & $0.79 \pm 0.01$ \\
DAG-DC-ADMM          & $1.00 \pm 0.00$ & $0.09 \pm 0.02$ & $0.87 \pm 0.03$ & $0.95 \pm 0.01$ & $0.26 \pm 0.03$ & $0.87 \pm 0.02$ \\
Oracle & $1.00 \pm 0.00$ & $0.07 \pm 0.02$ & $0.90 \pm 0.03$ & $0.89 \pm 0.03$ & $0.16 \pm 0.04$ & $0.93 \pm 0.02$ \\
\bottomrule
\end{tabular}
\end{table}

\subsubsection{Large-sample/Short-series scenario}
In this scenario, we use $n=200$ subjects with $m=50$ independent measurements and the same disturbance scales $\sigma\in\{0.5,1,2\}$. 
At $\sigma=0.5$, the model tends to over-segment (larger $n_{\text{est}}$), yet the resulting clusters remain internally coherent, as reflected by high homogeneity ($0.90$). Moreover, the near-equality of $\text{Recon.\ Error}_{\text{truth}}$ and $\text{Recon.\ Error}_{\text{est}}$ shows that, despite finer partitions, the learned labels deliver reconstruction quality comparable to that under the ground truth. As $\sigma$ increases to $1$ and $2$, over-segmentation is alleviated ($n_{\text{est}}=3$ and $5$), completeness rises, and ARI increases while homogeneity remains high ($> 0.90$). 
Reconstruction errors grow monotonically with $\sigma$ as expected, and remain nearly identical under true versus estimated labels across all $\sigma$ levels.

\begin{table}[h!]
\centering
\small
\caption{Clustering Results (Large-sample/Short-series, $n=200$, $m=50$). Values are mean $\pm$ 95\% CI over 50 runs; 
The true number of clusters $R_{\text{truth}}=2$ is fixed, and estimated $\widehat{R}$ is reported as the mode over repetitions.}
\setlength{\tabcolsep}{3pt} 
\begin{tabular}{lcccccccc}
\toprule
Setting & ARI & Homogeneity & Completeness & $\text{Recon.\ Error}_{\text{truth}}$ & $\text{Recon.\ Error}_{\text{est}}$ & $R_{\text{truth}}$ & $\widehat{R}$ \\
\midrule
$\sigma=0.5$ & $0.44 \pm 0.06$ & $0.90 \pm 0.07$ & $0.36 \pm 0.04$ & $1.25 \pm 0.00$ & $1.25 \pm 0.00$ & 2 & 18 \\
$\sigma=1$   & $0.68 \pm 0.10$ & $0.98 \pm 0.01$ & $0.64 \pm 0.09$ & $4.99 \pm 0.01$ & $4.99 \pm 0.01$ & 2 & 3 \\
$\sigma=2$   & $0.86 \pm 0.08$ & $0.92 \pm 0.07$ & $0.76 \pm 0.06$ & $19.96 \pm 0.03$ & $19.98 \pm 0.07$ & 2 & 5 \\
\bottomrule
\end{tabular}
\label{tab:clust_selected_metrics_N200_2C}
\end{table}

Across both the small-sample long-series and large-sample short-series regimes, and for all
$\sigma \in \{0.5,1,2\}$, the proposed DAG-DC-ADMM consistently outperforms the baselines
(Population and Individual) in structure recovery
(Tables~\ref{tab:combined_recovery_metrics_N50} and
\ref{tab:combined_recovery_metrics_N200}).
The improvement is consistent across both the undirected skeleton and directed DAG stages:
our method achieves higher TPR, lower FDR, and higher TNR than either baseline.
True-edge detection remains close to the oracle at both stages, and the small residual gap
is largely due to the model’s suppression of weak or ambiguous edges and minor orientation errors.
As $\sigma$ increases, TPR remains near-oracle while FDR approaches the oracle level,
demonstrating stable false-discovery control without sacrificing TPR.

These patterns can be understood from the interaction between cross-subject consistency and the consensus regularization in DAG--DC--ADMM.
Within each cluster, true causal dependencies tend to recur across subjects, whereas spurious edges tend to vary inconsistently. 
The consensus penalty promotes edges that are expressed consistently across subjects and shrinks those that appear inconsistently. Because $\mathbf{X}_i = \mathbf{Z}_i(\mathbf{I}-\mathbf{W}_i)^{-1}$,
increasing $\sigma$ scales all variances uniformly but leaves the causal relationships implied by $\mathbf{W}_i$ unchanged.
As a result, TPR remains high across $\sigma$ levels while FDR decreases and TNR increases as inconsistent edges are progressively suppressed.

\begin{table}[h]
\centering
\small 
\caption{Skeleton and DAG recovery metrics for different methods under varying disturbance levels ($\sigma$). (Large‐sample/Short‐series, $n=200,\;m=50$; mean $\pm$ 95\% CI over 50 runs, threshold $=0.02$).}
\label{tab:combined_recovery_metrics_N200}
\setlength{\tabcolsep}{3pt} 
\begin{tabular}{l|c|r|r|r|r|r}
\toprule
Method & Skel TPR & Skel FDR & Skel TNR & DAG TPR & DAG FDR & DAG TNR \\
\midrule
\multicolumn{7}{l}{\textbf{Disturbance Scale $\sigma = 0.5$}} \\
\midrule
Population  & $0.93 \pm 0.03$ & $0.46 \pm 0.02$ & $0.18 \pm 0.05$ & $0.64 \pm 0.04$ & $0.63 \pm 0.02$ & $0.63 \pm 0.02$ \\
Individual  & $0.95 \pm 0.00$ & $0.39 \pm 0.00$ & $0.36 \pm 0.00$ & $0.66 \pm 0.01$ & $0.58 \pm 0.01$ & $0.69 \pm 0.00$ \\
DAG-DC-ADMM          & $0.99 \pm 0.01$ & $0.21 \pm 0.02$ & $0.68 \pm 0.04$ & $0.87 \pm 0.02$ & $0.38 \pm 0.03$ & $0.79 \pm 0.02$ \\
Oracle & $1.00 \pm 0.00$ & $0.08 \pm 0.02$ & $0.89 \pm 0.03$ & $0.88 \pm 0.03$ & $0.18 \pm 0.04$ & $0.92 \pm 0.02$ \\
\midrule
\multicolumn{7}{l}{\textbf{Disturbance Scale $\sigma = 1.0$}} \\
\midrule
Population  & $0.93 \pm 0.03$ & $0.46 \pm 0.02$ & $0.19 \pm 0.04$ & $0.64 \pm 0.04$ & $0.63 \pm 0.03$ & $0.63 \pm 0.02$ \\
Individual  & $0.88 \pm 0.01$ & $0.28 \pm 0.00$ & $0.63 \pm 0.00$ & $0.61 \pm 0.01$ & $0.51 \pm 0.01$ & $0.79 \pm 0.00$ \\
DAG-DC-ADMM          & $0.99 \pm 0.00$ & $0.11 \pm 0.02$ & $0.85 \pm 0.03$ & $0.87 \pm 0.02$ & $0.25 \pm 0.04$ & $0.88 \pm 0.02$ \\
Oracle & $1.00 \pm 0.00$ & $0.07 \pm 0.02$ & $0.91 \pm 0.03$ & $0.88 \pm 0.03$ & $0.17 \pm 0.04$ & $0.93 \pm 0.02$ \\
\midrule
\multicolumn{7}{l}{\textbf{Disturbance Scale $\sigma = 2.0$}} \\
\midrule
Population & $0.94 \pm 0.03$ & $0.46 \pm 0.02$ & $0.19 \pm 0.05$ & $0.64 \pm 0.04$ & $0.63 \pm 0.02$ & $0.63 \pm 0.02$ \\
Individual & $0.96 \pm 0.00$ & $0.43 \pm 0.00$ & $0.27 \pm 0.00$ & $0.66 \pm 0.01$ & $0.60 \pm 0.01$ & $0.66 \pm 0.00$ \\
DAG-DC-ADMM & $0.99 \pm 0.01$ & $0.07 \pm 0.03$ & $0.89 \pm 0.04$ & $0.89 \pm 0.03$ & $0.16 \pm 0.04$ & $0.93 \pm 0.02$ \\
Oracle & $1.00 \pm 0.00$ & $0.07 \pm 0.02$ & $0.90 \pm 0.03$ & $0.89 \pm 0.03$ & $0.16 \pm 0.04$ & $0.93 \pm 0.02$ \\
\bottomrule
\end{tabular}
\end{table}

In the large-sample short-series regime, the consensus effect intensifies as the number of subjects grows.
With more subjects per cluster, the statistical signal strengthens: true dependencies are consistently observed across subjects, while subject-specific artifacts behave as outliers penalized by the consensus term.
This yields tighter false-discovery control and higher TNR without diminishing recall,
further widening the performance gap between DAG-DC-ADMM and the NOTEARS baselines. The Supplementary Material reports additional results, including a three-cluster scenario and thresholding robustness analysis where we sweep the $\delta$ to evaluate robustness of skeleton and directed DAG recovery across different sparsity levels.

\section{Case Study}\label{s:case_study}
We evaluate our method on the multivariate single-cell dataset of \citet{sachs2005causal}, which consists of flow cytometry measurements from thousands of primary human $\text{CD4}^+\text{T}$ cells collected under nine perturbation conditions. Cell reactions are fixed 15 minutes after stimulation to capture how each perturbation affects the intracellular signaling networks. For each cell, the amounts of 11 molecules—Raf, Mek, Erk, P38, Jnk, Akt, PKC, PKA, PLC-$\gamma$, and the phospholipids PIP$_2$ and PIP$_3$—are measured. 

\subsection{Data Preprocessing}\label{subsec:preprocessing}

In the \citet{sachs2005causal} dataset, approximately 7{,}745 single-cell measurements are collected and split into nine experimental subsets, each corresponding to a distinct biochemical perturbation condition. Single-cell signaling responses are inherently heterogeneous, even under the same external perturbation, and signal loss can occur when unique response patterns of heterogeneous cellular subsets are masked by averaging techniques \citep{sachs2005causal}. To address this and recover meaningful structure, we aggregate cells within each perturbation into coherent subpopulations via a pre-clustering step. 

For each perturbation dataset, we first Z-score standardize all 11 signaling features. We then apply Principal Component Analysis (PCA) \citep{jolliffe2011principal} and retain the smallest number of principal components that explain at least $80\%$ of the total variance. These low-dimensional PCA embeddings then serve as input for $K$-means clustering. To determine the optimal number of subpopulations, we run $K$-means between two and ten clusters and evaluate the results using the Silhouette Score \citep{shahapure2020cluster}. We then choose the smallest number of clusters whose Silhouette Score lies within $1\%$ of the maximum value. Finally, to ensure robustness of subsequent causal dependency structure estimation, any subpopulation containing fewer than 20 cells is removed from the analysis. This procedure yields a total of 28 distinct subpopulations across the nine perturbation conditions, which we use as input to the DAG-DC-ADMM algorithm.

\subsection{Results}\label{subsec:results}
We consider three modeling scenarios with different assumptions. First, a pooled-population model that assumes all cells share a single response pattern across perturbations. This yields one causal dependency structure for the entire dataset. Second, a per-perturbation model, which assumes that cells respond similarly within each perturbation but may differ across perturbations, resulting in nine causal dependency structures. Third, our method, which takes the 28 subpopulations as the input. It identifies 6 clusters, each with its own causal dependency structure. For all scenarios, we tune hyperparameters via the same three-fold cross-validation. For the first two scenarios, we tune the sparsity parameter $\lambda_{1}$ over $\{0.001, 0.005, 0.01, 0.1\}$. For the DAG-DC-ADMM algorithm, we search over the following grid:
$
\lambda_{1} \in \{0.001, 0.005, 0.01, 0.1\}$,
$\lambda_{2} \in \{0.0001, 0.001,0.005,0.01,0.03,0.05,0.1\}$,
$\tau \in \{0.7, 1, 1.5, 2, 2.5, 3, 3.5, 4\},
$
while fixing $\rho_{2}=0.05$ and initializing $\rho_{1}=0.1$.
The optimal combination of parameters is selected by minimizing the mean reconstruction error on the validation folds.

Table~\ref{tab:recon_comparison} reports the resulting average validation reconstruction errors for these three scenarios. The results show that DAG-DC-ADMM yields a smaller reconstruction error, outperforming both the pooled population model and the per-perturbation model. This observation bolsters the hypothesis that cells respond differently even to the same perturbation, and grouping by perturbation might not be the best approach for understanding signaling pathways. 

\begin{table}[ht]
\centering
\small
\caption{Reconstruction error comparison under three-fold cross-validation.}
\label{tab:recon_comparison}
\setlength{\tabcolsep}{3pt}
\begin{tabular}{lcc}
\toprule
Modeling scenario & Fitting granularity & Average Validation Recon. Error \\
\midrule
Pooled population & 1 population-level model (7{,}446 cells) & 9.02 \\
Per-perturbation   & 9 perturbation-level models & 8.66 \\
DAG-DC-ADMM & 6 cluster-level models & 8.50 \\
\bottomrule
\end{tabular}
\end{table}

\begin{figure}[H]
    \centering
    \begin{subfigure}[b]{0.32\linewidth}
        \centering
        \includegraphics[width=\linewidth]{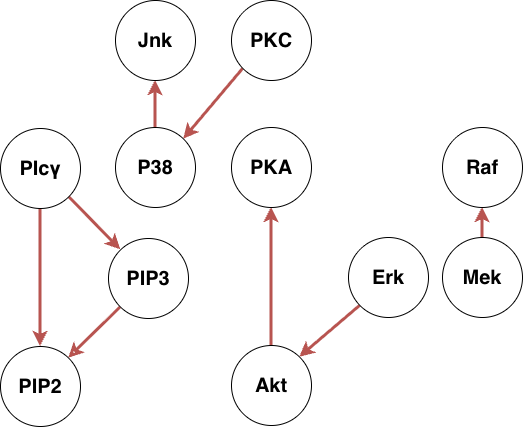}
        \caption{Population-level DAG}
        \label{fig:pop_dag}
    \end{subfigure}
    \hfill
    \begin{subfigure}[b]{0.32\linewidth}
        \centering
        \includegraphics[width=\linewidth]{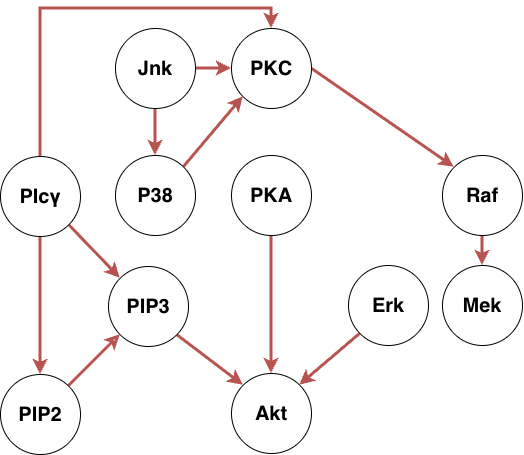}
        \caption{Cluster 2 DAG}
        \label{fig:cluster2_dag}
    \end{subfigure}
    \hfill
    \begin{subfigure}[b]{0.32\linewidth}
        \centering
        \includegraphics[width=\linewidth]{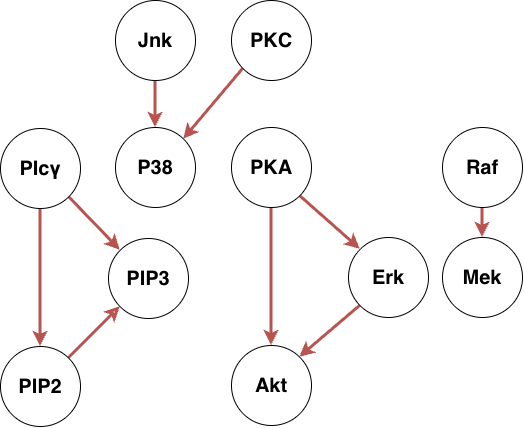}
        \caption{Cluster 4 DAG}
        \label{fig:cluster4_dag}
    \end{subfigure}
    \caption{Comparison of DAG structures across population and representative clusters.}
    \label{fig:dag_comparison}
\end{figure}

\begin{figure}[H]
    \centering
\includegraphics[width=0.95\linewidth]{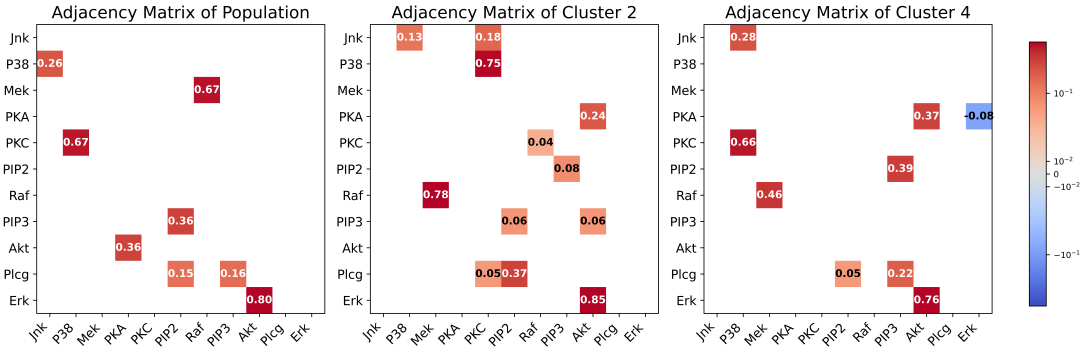}
    \caption{Adjacency matrix comparison}
    \label{fig:adjacency_matrix_comparison}
\end{figure}

On heterogeneous causal graph discovery, our method identified 6 distinct clusters. Sizes range from small cohorts (e.g., cluster 1: 99 cells, cluster 6: 24 cells) to larger groups (e.g., cluster 4: 4,117 cells). Each cluster presents a distinct adjacency matrix. They differ in edge presence (or absence), and in the sign and magnitude of edge weights—differences that a single global model would obscure. Fig. \ref{fig:dag_comparison} and Fig. \ref{fig:adjacency_matrix_comparison} present the DAGs and adjacency matrices of the global model, and two randomly picked clusters, to illustrate these contrasts. For example, Cluster 2 contains directed links that are absent in Cluster 4 and in the population graph, such as edges from PIP3 to Akt and PLC-$\gamma$ to PKC. Biologically, the edge from PIP3 to Akt reflects the PI3K-Akt signaling axis, where PIP3 serves as a membrane anchor that recruits Akt for phosphorylation‐based activation \citep{hoxhaj2020pi3k}. The edge from PLC-$\gamma$ to PKC represents activation of PKC through the PLC-$\gamma$ pathway. PLC-$\gamma$ hydrolyzes the membrane lipid PIP2 into two second messengers: diacylglycerol and IP3. Diacylglycerol remains in the membrane and directly activates PKC \citep{nosbisch2020mechanistic}.
The selective presence of these edges across clusters indicates subpopulation heterogeneity. In some subpopulations, these pathways appear coherently engaged, in line with stronger upstream receptor signaling and more favorable cofactor conditions. In others, the connections are weaker, diverted through alternative routes, or suppressed, for example through dephosphorylation by phosphatases or competition from parallel signaling pathways.

In contrast, the population-level DAG, built by pooling all cells, omits connections that are clearly present within specific subpopulations. This is a classic effect of population averaging: heterogeneous signaling states are smoothed together, and edges that are strong only in certain subsets fall below detection \citep{sachs2005causal}. The identification of these biologically meaningful links confirms that distinct cellular reaction patterns coexist in the data, emphasizing the importance of a unified clustering and graph structure learning framework that is essential to recover specific causal dependency mechanisms that a single global model would mask.

\section{Discussion and Conclusion}\label{s:discussion}
This work presents DAG–DC–ADMM, a unified framework for joint causal structure learning and structure-aware clustering. Our method recovers cluster-specific causal dependency structures from heterogeneous data and achieves superior structure recovery accuracy compared with baseline methods. The framework provides a foundation for modeling population heterogeneity in underlying causal mechanisms. We next outline future directions: (i) integrating temporal causal relationships, (ii) extending to nonlinear dependencies with nonparametric SEMs, {and (iii) allowing heterogeneous exogenous disturbance distributions across subjects and formalizing the associated clustering criterion.}

Our framework naturally extends to dynamic settings where time-varying behaviors are captured using time-series data. Specifically, for each subject $i$, the data matrix $\mathbf{X}_i$ represents the measurements of $d$ features over $m_i$ time points, where each row corresponds to a time index. With such data, we can formulate a dynamic SEM that incorporates lagged effects: $
\mathbf{X}_i = \mathbf{X}_i \mathbf{W}_i + \sum_{q=1}^{p} \mathbf{Y}_{i,q} \mathbf{A}_{i,q} + \mathbf{Z}_i,$
where $\mathbf{W}_i$ captures contemporaneous dependencies, $\mathbf{Y}_{i,q} \in \mathbb{R}^{m \times d}$ stores the lagged measurements for subpopulation $i$ at lag $q$, and $\{\mathbf{A}_{i,q}\}_{q=1}^{p}$ encode lagged dependencies for subject $i$. This extension allows for the joint estimation of both intra-time and inter-time dependencies with acyclicity enforced on $\mathbf{W}_i$. Crucially, our existing consensus penalty can be readily adapted to jointly regularize both contemporaneous $\mathbf{W}_i$ and lagged $\{\mathbf{A}_{i,q}\}_{q=1}^{p}$ dependency matrices, thereby identifying structure-aware clusters based on dynamic causal dependencies.

Another promising future direction is to incorporate nonlinear dependencies through nonparametric SEMs, in which each feature is generated by a differentiable function $f$ of its parent features plus independent unexplained variability. Following recent advances in nonparametric DAG learning \citep{zheng2020learning}, the edge strength could be defined through functional derivatives, allowing flexible modeling of causal effects beyond linearity. The core of this extension involves minimizing a differentiable loss function $L(f)$ over the functional space, subject to a continuous acyclicity constraint expressed through the trace exponential as $h(\mathbf{W}(f))=0$. Developing an effective way to integrate this nonlinear optimization problem into the clustering framework and define meaningful structure similarity metrics to facilitate clustering efficiency remains a challenge for future research.

\if0\blind{
\section*{Acknowledgements}
The authors acknowledge the support from the funding agency (removed for double-blinded review).	} \fi

\bibliographystyle{apalike}
\spacingset{1}
\bibliography{IISE-Trans}

\newpage

\input{IISE_Transactions_Appendix.tex}
\end{document}

%% file: IISE_Transactions_Appendix.tex
\section*{Supplementary Material for Manuscript ``A Unified Framework for Structure Aware Clustering and Heterogeneous Causal Graph Learning"}\label{s:Appendix}

\setcounter{section}{0}
\setcounter{subsection}{0}
\setcounter{figure}{0}
\setcounter{table}{0}
\setcounter{equation}{0}

\renewcommand{\thefigure}{S\arabic{figure}}
\renewcommand{\thetable}{S\arabic{table}}
\renewcommand{\theequation}{S\arabic{equation}}
\renewcommand{\thesection}{\Alph{section}}

\section{Proofs of Theorems~\ref{thm:convergence} and~\ref{thm:finite}}
\paragraph{Proof of Theorem \ref{thm:convergence}.}
Each term in $\mathcal{L}_{\mathrm{aug}}^{(s)}$ is either a smooth semi–algebraic function (quadratic loss, acyclicity penalty) or a proper, lower–semicontinuous semi–algebraic function ($\ell_1$, gTLP). Therefore, $\mathcal{L}_{\mathrm{aug}}^{(s)}$ is proper, lower–semicontinuous (LSC), and satisfies the KL property \citep{attouch2010proximal,attouch2013convergence}. The coercivity of the quadratic terms in the full Lagrangian implies that the ADMM sequence $\{(\mathbf{W}^{k},\boldsymbol{\theta}^{k})\}$ is bounded. 

The updates of $\mathbf W$ and $\boldsymbol\theta$ are obtained via sequential minimization:$$\mathbf W^{k+1} = \arg\min_{\mathbf W}\; \mathcal{L}_{\mathrm{aug}}^{(s)}(\mathbf W,\boldsymbol\theta^{k}), \quad \boldsymbol\theta^{k+1} = \arg\min_{\boldsymbol\theta}\; \mathcal{L}_{\mathrm{aug}}^{(s)}(\mathbf W^{k+1},\boldsymbol\theta).$$

The reconstruction loss and acyclicity terms in $\mathcal{L}_{\mathrm{aug}}^{(s)}$ have Lipschitz-continuous gradients on bounded sets, while the $\ell_1$ and gTLP terms are proper
lower-semicontinuous with bounded subgradients on bounded sets. In addition, the quadratic term $\rho_2\sum_{i<j}\bigl\| \boldsymbol{\theta}_{ij}-(\mathbf W_i-\mathbf W_j) \bigr\|_F^2/2$ is strongly convex jointly in $(\mathbf W,\boldsymbol\theta)$ and is proportional to $\rho_2$. Standard non-convex ADMM theory confirms that because of this strong convexity and the Lipschitz continuity, the sequential minimization satisfies the sufficient decrease condition (H1).

More explicitly, for the $\mathbf W$-step, the strong convexity guarantees the existence of a constant $g_1>0$ such that:$$\mathcal{L}_{\mathrm{aug}}^{(s)}(\mathbf W^{k+1},\boldsymbol\theta^{k}) \;\le\; \mathcal{L}_{\mathrm{aug}}^{(s)}(\mathbf W^{k},\boldsymbol\theta^{k}) - g_1 \bigl\|\mathbf W^{k+1}-\mathbf W^{k}\bigr\|_F^2.$$Similarly, the $\boldsymbol\theta$-step yields a constant $g_2>0$ such that:$$\mathcal{L}_{\mathrm{aug}}^{(s)}(\mathbf W^{k+1},\boldsymbol\theta^{k+1}) \;\le\; \mathcal{L}_{\mathrm{aug}}^{(s)}(\mathbf W^{k+1},\boldsymbol\theta^{k}) - g_2 \bigl\|\boldsymbol\theta^{k+1}-\boldsymbol\theta^{k}\bigr\|_F^2.$$By summing these two inequalities and setting $a=\min(g_1, g_2) > 0$, we obtain the overall sufficient decrease condition (H1) for the sequence $\{(\mathbf{W}^k, \boldsymbol{\theta}^k)\}$:
$$\mathcal{L}_{\mathrm{aug}}^{(s)}(\mathbf W^{k+1},\boldsymbol\theta^{k+1})\le\mathcal{L}_{\mathrm{aug}}^{(s)}(\mathbf W^{k},\boldsymbol\theta^{k})-a\Bigl( \bigl\|\mathbf W^{k+1}-\mathbf W^{k}\bigr\|_F^2 + \bigl\|\boldsymbol\theta^{k+1}-\boldsymbol\theta^{k}\bigr\|_F^2 \Bigr).$$

Moreover, the ADMM dual update for the consensus constraint is $\mathbf{U}_{ij}^{k+1} = \mathbf{U}_{ij}^{k} + \boldsymbol{\theta}_{ij}^{k+1} - (\mathbf{W}_i^{k+1} - \mathbf{W}_j^{k+1})$. This implies that the constraint residual is proportional to the change in the dual variables: $\boldsymbol{\theta}_{ij}^{k+1} - (\mathbf{W}_i^{k+1} - \mathbf{W}_j^{k+1}) = \mathbf{U}_{ij}^{k+1} - \mathbf{U}_{ij}^{k}$. The first-order optimality conditions for the $\mathbf{W}$- and $\boldsymbol{\theta}$-updates then give a subgradient $\boldsymbol{\omega}^{k+1}
\in
\partial \mathcal{L}_{\mathrm{aug}}^{(s)}(\mathbf{W}^{k+1}, \boldsymbol{\theta}^{k+1})
$. It combines (i) gradients of smooth terms with Lipschitz-continuous gradient, (ii) bounded subgradients of the nonsmooth penalties on bounded level sets, and (iii) the quadratic consensus term, which is proportional to the dual step $\mathbf{U}_{ij}^{k+1}-\mathbf{U}_{ij}^{k}$.

Consequently, there exists a constant $L>0$, depending only on the Lipschitz constants of the smooth terms and the fixed penalty parameter $\rho_2$, such that the relative error condition (H2) holds:

$$\|\boldsymbol{\omega}^{k+1}\|_F \le L \left( \|\mathbf{W}^{k+1} - \mathbf{W}^{k}\|_F + \|\boldsymbol{\theta}^{k+1} - \boldsymbol{\theta}^{k}\|_F \right).$$

For continuity condition (H3), the ADMM sequence $\{(\mathbf W^{k}, \boldsymbol{\theta}^{k})\}$ is bounded, thus guaranteeing the existence of a convergent subsequence. And the objective function $\mathcal{L}_{\text{aug}}^{(s)}(\mathbf W, \boldsymbol{\theta})$ is lower-semicontinuous. By the sufficient decrease condition (H1) and the boundedness of the sequence, the objective values $\mathcal{L}_{\text{aug}}^{(s)}(\mathbf W^{k}, \boldsymbol{\theta}^{k})$ form a non-increasing sequence. Since the objective function is bounded below, this sequence of objective values is guaranteed to converge to a finite limit.

The LSC property of $\mathcal{L}_{\text{aug}}^{(s)}$ ensures that when the sequence converges to the limit point $(\mathbf W^{\star}, \boldsymbol{\theta}^{\star})$, the function values satisfy:
$$\mathcal{L}_{\text{aug}}^{(s)}(\mathbf W^{\star}, \boldsymbol{\theta}^{\star}) \le \liminf_{k \to \infty} \mathcal{L}_{\text{aug}}^{(s)}(\mathbf W^{k}, \boldsymbol{\theta}^{k}).$$

This convergence of the objective function values to the function value at the limit point verifies continuity condition (H3).

Because the augmented objective $\mathcal{L}_{\mathrm{aug}}^{(s)}$ satisfies these conditions, the convergence result of Attouch et al. applies. Therefore, by Lemma 2.6 and Theorem 2.9 in \citet{attouch2013convergence}, the entire sequence $\{(\mathbf{W}^k, \boldsymbol{\theta}^k)\}$ has finite length and converges to a first-order stationary point $(\mathbf{W}^{\star}, \boldsymbol{\theta}^{\star})$ of the Augmented Lagrangian $\mathcal{L}_{\mathrm{aug}}^{(s)}$.

\paragraph{Proof of Theorem~\ref{thm:finite}.}
The convergence of the outer difference-of-convex iteration sequence $\{(\mathbf W^{(s)},\boldsymbol{\theta}^{(s)})\}_{s\ge 0}$ to a KKT point in a finite number of steps is guaranteed by the fundamental properties of the difference-of-convex programming framework, as leveraged in prior work on penalized clustering \citep{wu2016new}. First, the convex majorization applied at each step ensures that the objective value $\mathcal F(\widehat{\mathbf W}^{(s)},\widehat{\boldsymbol{\theta}}^{(s)})$ is monotonically non-increasing throughout the optimization, as the convex objective $\mathcal F^{(s)}(\cdot)$ is constructed to upper-bound the true objective $\mathcal F(\cdot)$ at the previous iteration. The monotonicity follows directly from the sequence of inequalities:
$$\begin{aligned}
0 \le \mathcal F\!\bigl(\widehat{\mathbf W}^{(s)},\widehat{\boldsymbol{\theta}}^{(s)}\bigr) &= \mathcal F^{(s+1)}\!\bigl(\widehat{\mathbf W}^{(s)},\widehat{\boldsymbol{\theta}}^{(s)}\bigr) \le \mathcal F^{(s)}\!\bigl(\widehat{\mathbf W}^{(s)},\widehat{\boldsymbol{\theta}}^{(s)}\bigr) \\
&\le \mathcal F^{(s)}\!\bigl(\widehat{\mathbf W}^{(s-1)},\widehat{\boldsymbol{\theta}}^{(s-1)}\bigr) = \mathcal F\!\bigl(\widehat{\mathbf W}^{(s-1)},\widehat{\boldsymbol{\theta}}^{(s-1)}\bigr).
\end{aligned}$$

Since the objective is bounded below, this sequence of objective values is guaranteed to converge. Second, and critically, the convex approximation objective $\mathcal F^{(s)}(\mathbf W,\boldsymbol{\theta})$ depends on the iteration $s$ only through the set of indicator functions $I(\|\boldsymbol{\theta}_{ij}^{(s-1)}\|_F < \tau)$. Because these indicators only take a binary value 0 or 1 for each of the $n(n-1)/2$ pairs, the function $\mathcal F^{(s)}(\mathbf W,\boldsymbol{\theta})$ can only adopt a finite number of distinct functional forms across all iterations. The combination of monotonic decrease and a finite set of possible optimal values guarantees that the algorithm must stop at a fixed point, $(\mathbf W^{(s^\star)},\boldsymbol{\theta}^{(s^\star)})$, within a finite number of iterations, $s^\star < \infty$. Standard difference-of-convex programming theory then asserts that this limit point is a KKT point of {$\mathcal F(\mathbf W,\boldsymbol{\theta})$}.

\section{Thresholding Robustness Analysis}
\begin{figure}[H]
    \centering
    \includegraphics[width=0.9\linewidth]{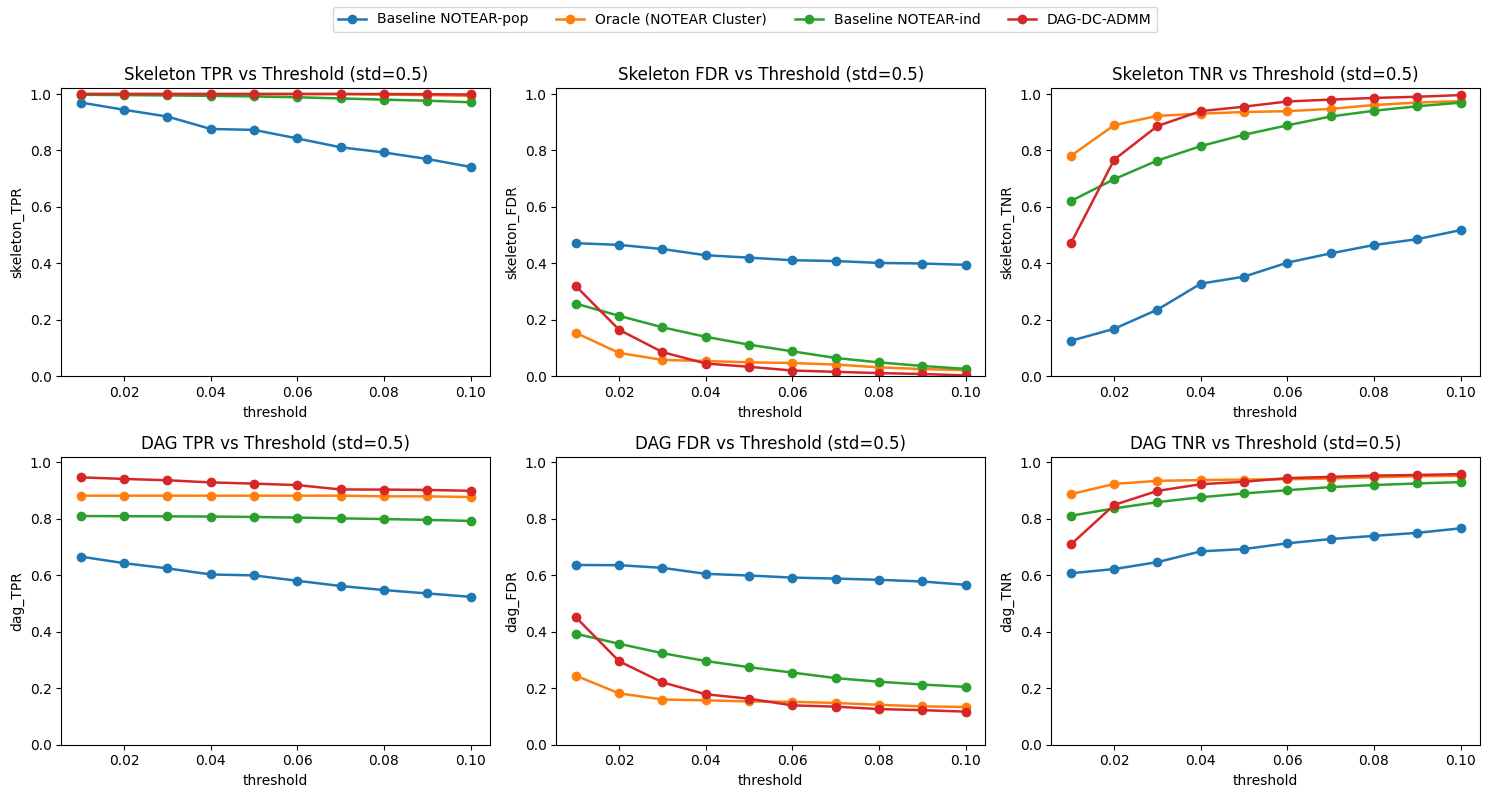}
    \caption{Thresholding robustness in the small-sample/long-series setting ($n=50$, $\sigma=0.5$).}
    \label{fig:N50std05thresholdplot}
\end{figure}

\begin{figure}[H]
    \centering
    \includegraphics[width=0.9\linewidth]{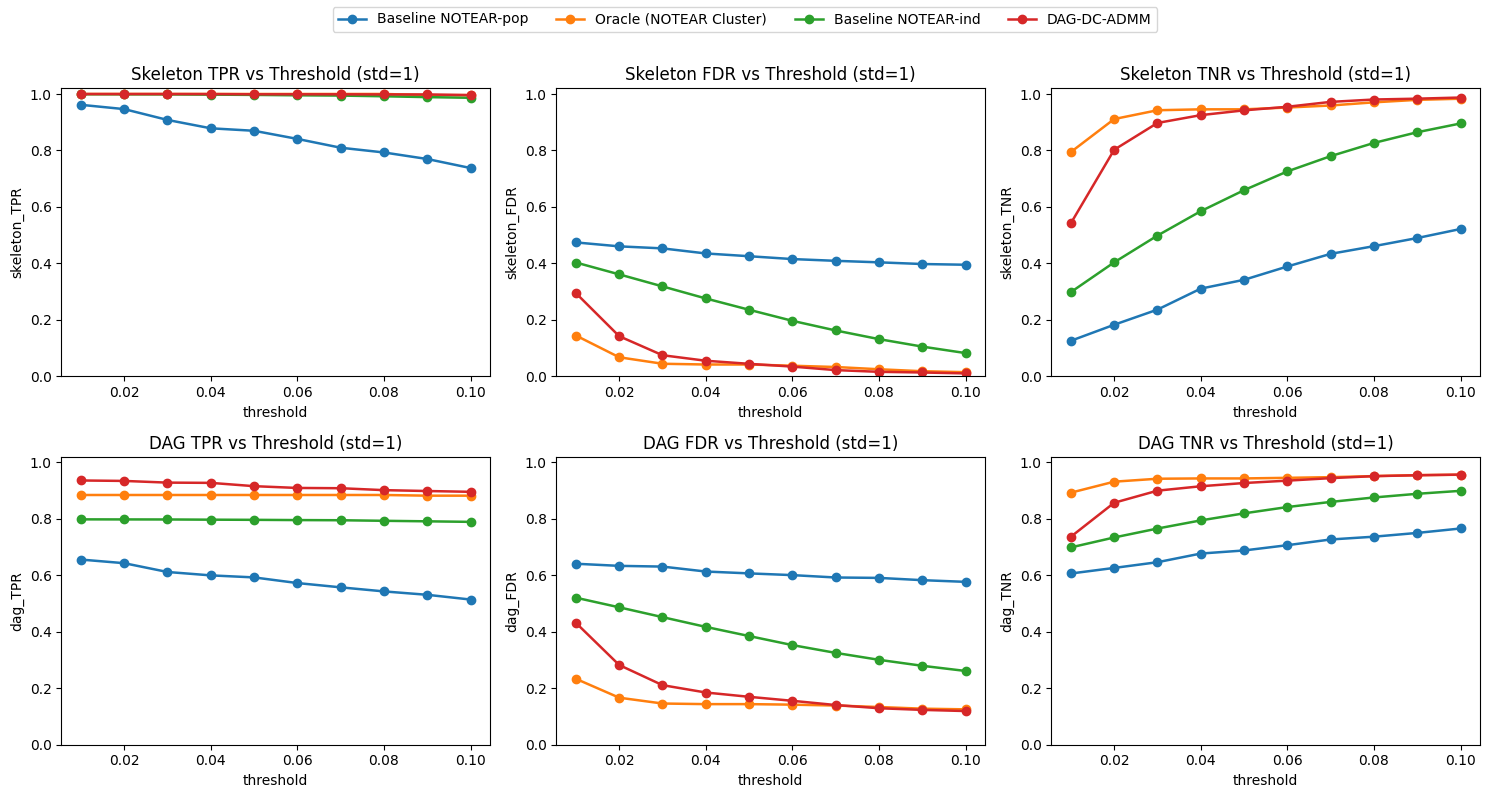}
    \caption{Thresholding robustness in the small-sample/long-series setting ($n=50$, $\sigma=1$).}
    \label{fig:N50std1thresholdplot}
\end{figure}

\begin{figure}[H]
    \centering
    \includegraphics[width=0.9\linewidth]{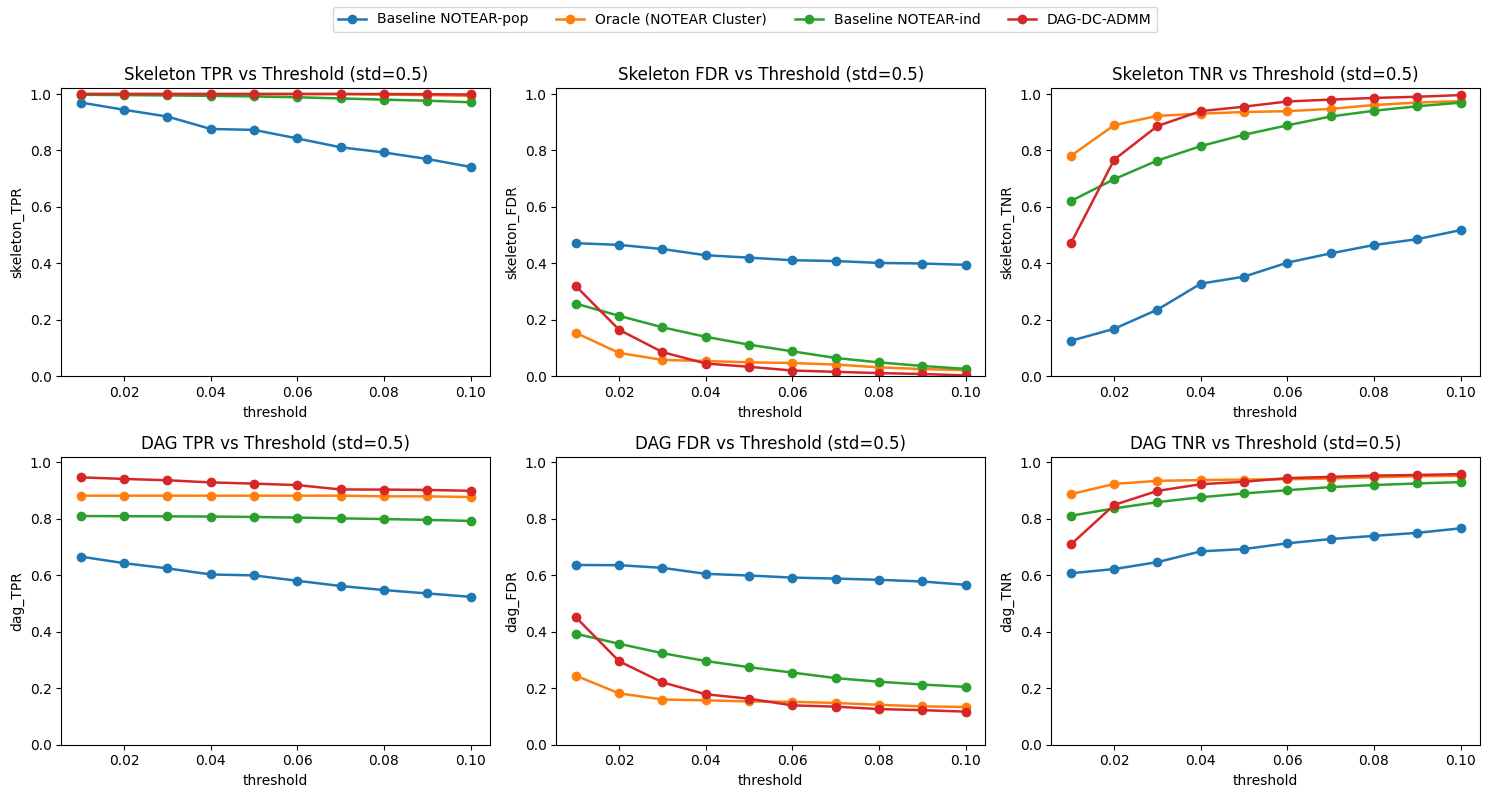}
    \caption{Thresholding robustness in the small-sample/long-series setting ($n=50$, $\sigma=2$).}
\label{fig:N50std2thresholdplot}
\end{figure}

\begin{figure}[H]
    \centering
    \includegraphics[width=0.9\linewidth]{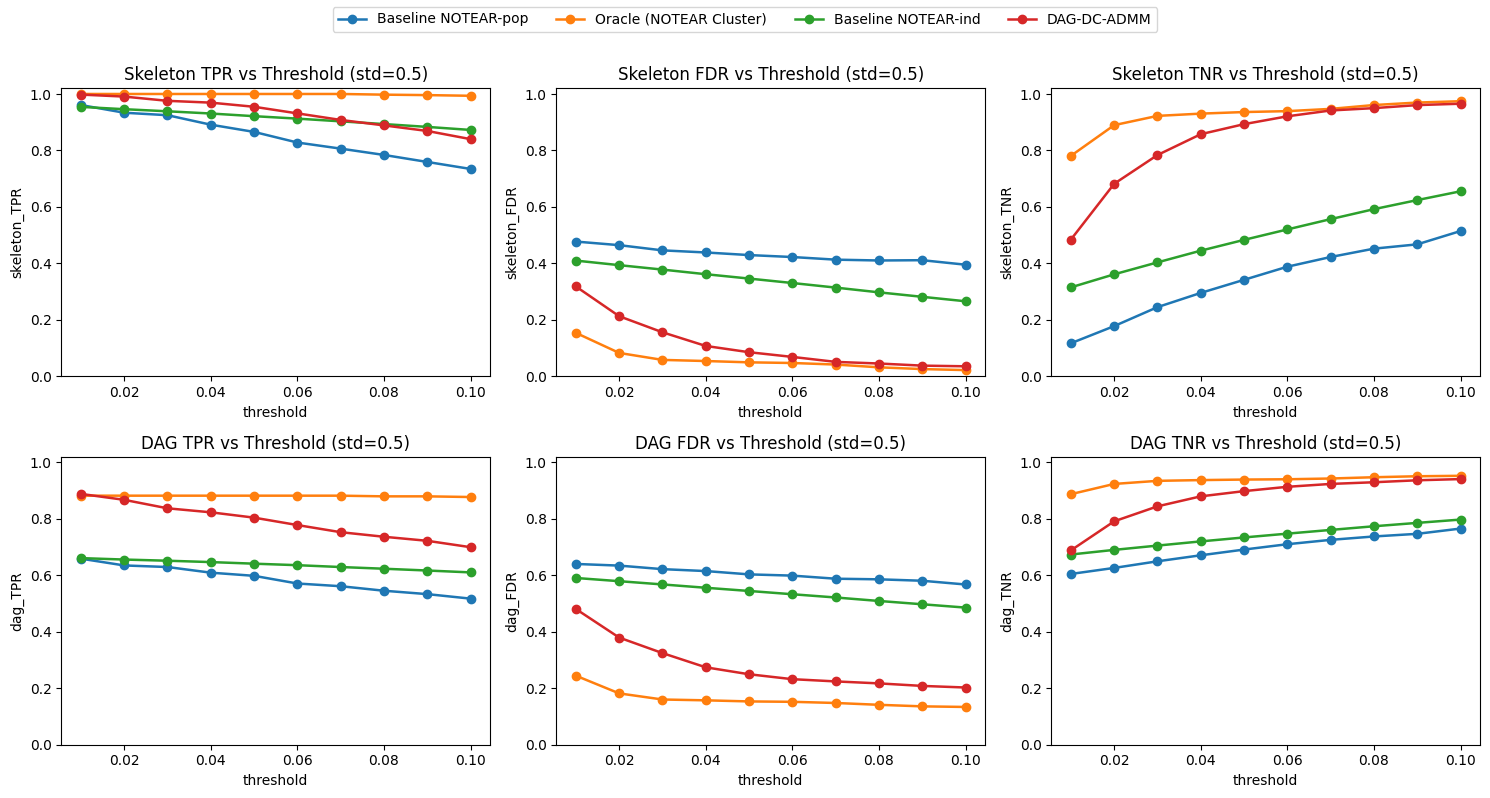}
\caption{Thresholding robustness in the small-sample/long-series setting ($n=200$, $\sigma=0.5$).}
\label{fig:N200std05thresholdplot}
\end{figure}

\begin{figure}[H]
    \centering
    \includegraphics[width=0.9\linewidth]{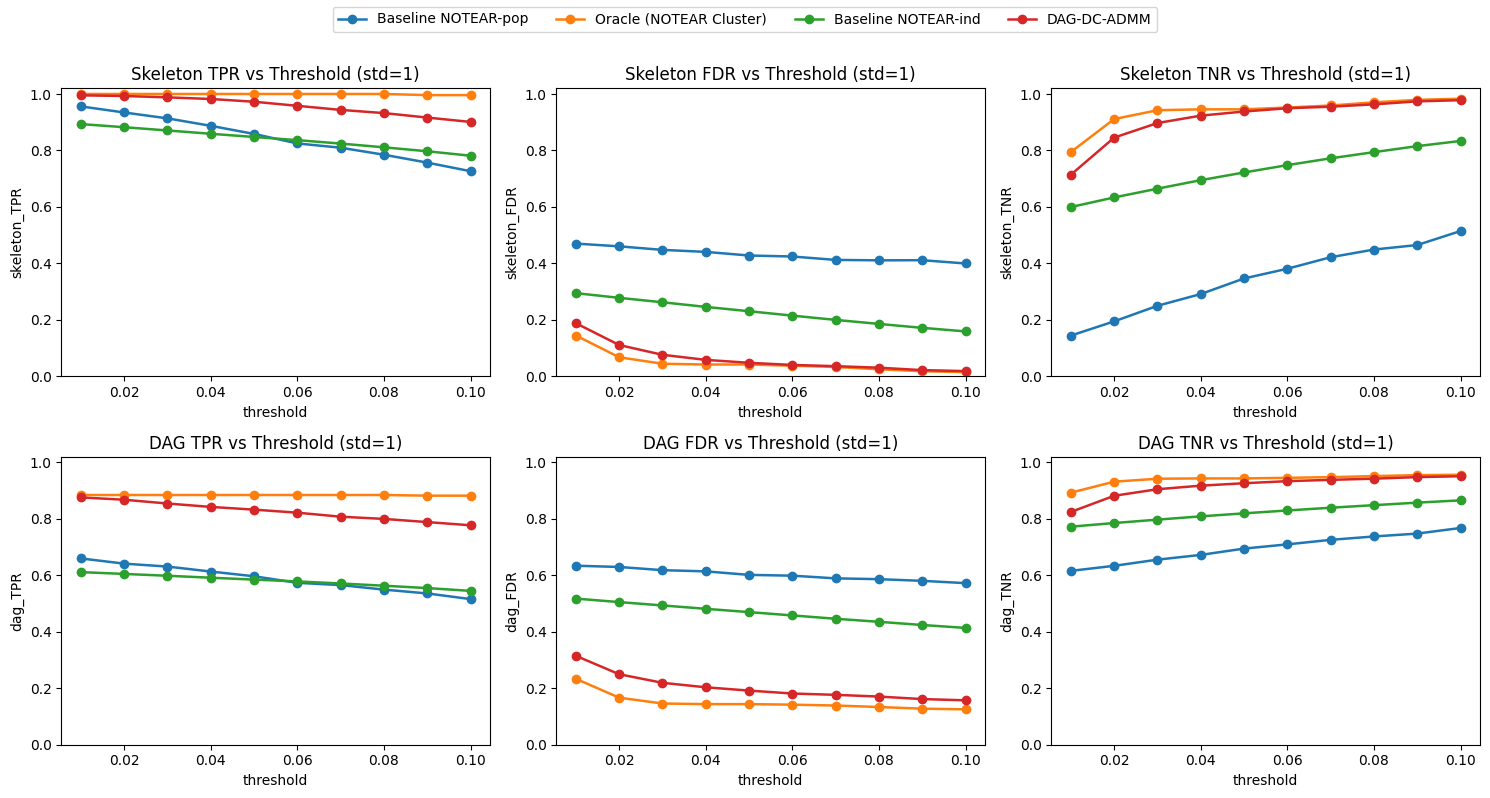}
    \caption{Thresholding robustness in the large-sample/short-series setting ($n=200$, $\sigma=1$).}
    \label{fig:N200std1thresholdplot}
\end{figure}

\begin{figure}[H]
    \centering
    \includegraphics[width=0.9\linewidth]{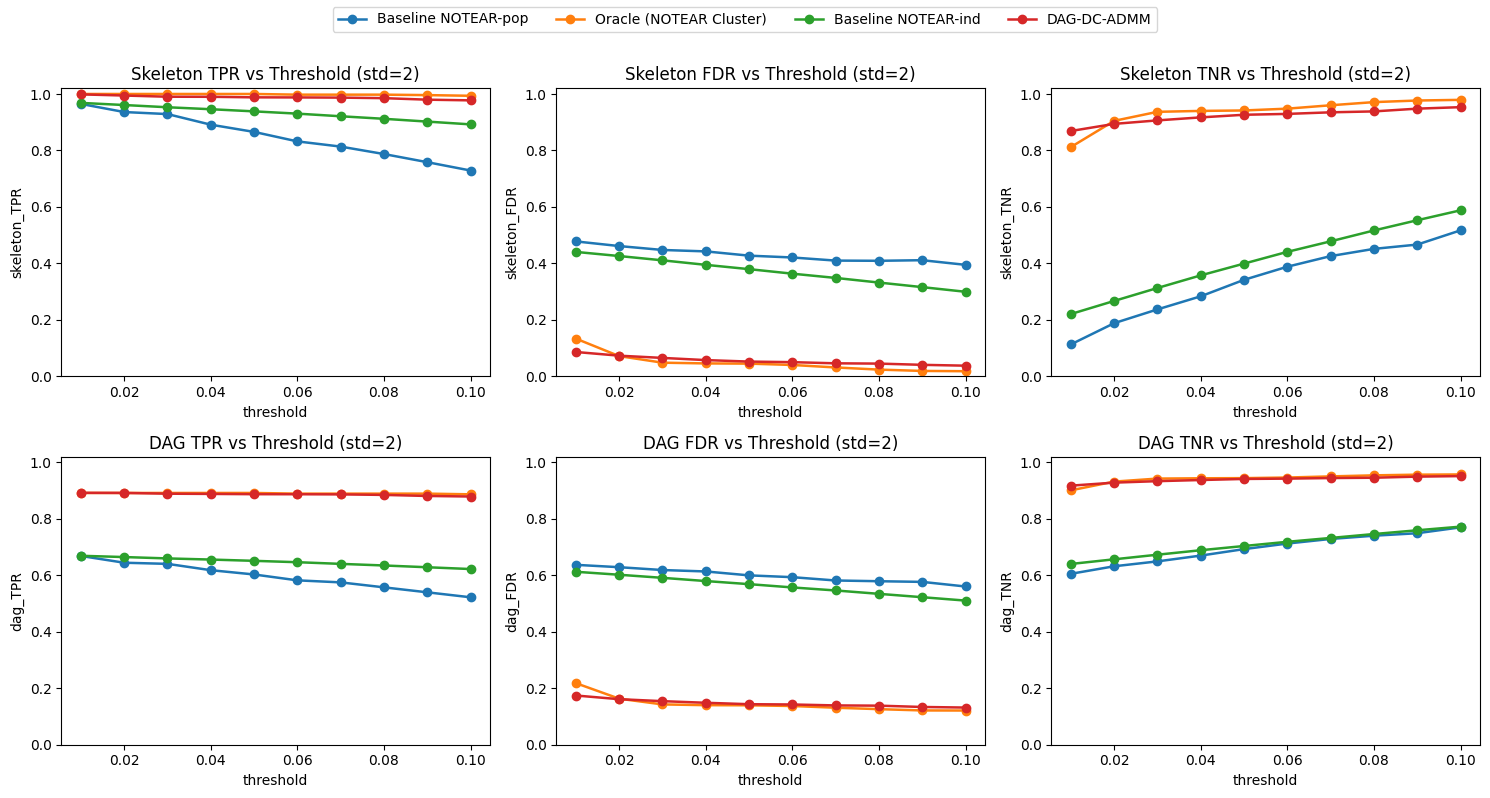}
    \caption{Thresholding robustness in the large-sample/short-series setting ($n=200$, $\sigma=2$).}
\label{fig:N200std2thresholdplot}
\end{figure}

In Fig.~\ref{fig:N50std05thresholdplot} to Fig.~\ref{fig:N200std2thresholdplot}, each panel reports a metric as the sparsity threshold $\delta$ varies from 0.01 to 0.10. Across the full threshold sweep, DAG-DC-ADMM consistently outperforms NOTEARS-Individual and NOTEARS-Population for both skeleton and DAG recovery. It achieves higher TPR together with lower FDR and higher TNR than both baselines. As the sparsity threshold $\delta$ increases, TPR declines slightly but still stays closer to the oracle than either baseline. Meanwhile, FDR decreases and TNR increases, and both metrics approach the oracle. This reflects stronger error control for only a modest trade-off in TPR. Moreover, there is a practical operating window $\delta \in [0.01, 0.04]$ where increasing $\delta$ improves FDR and TNR without materially affecting TPR, rendering conclusions insensitive to the precise threshold choice. With respect to threshold selection, the method requires minimal tuning: it tends to separate true from spurious edges with a clear margin, so tightening the threshold mainly removes spurious edges while leaving true ones largely intact.

\section{Three-Cluster Scenario}
We additionally evaluated a three-cluster scenario for both the small-sample/long-series case and the large-sample/short-series case; the same conclusions hold. This confirms that our method's advantages over baselines persist under multi-cluster scenario.

\begin{table}[ht]
\centering
\small
\caption{Three Clusters: Clustering Results (Small-sample/Long-series, $n=50$, $m=300$). Values are mean $\pm$ 95\% CI over 50 runs; 
The true number of clusters $R_{\text{truth}}=3$ is fixed, and estimated $\widehat{R}$ is reported as the mode over repetitions.}
\setlength{\tabcolsep}{3pt} 
\begin{tabular}{lccccccc}
\toprule
Setting & ARI & Homogeneity & Completeness & $\text{Recon.\ Error}_{\text{truth}}$& $\text{Recon.\ Error}_{\text{est}}$ & $R_{\text{truth}}$ & $\widehat{R}$\\
\midrule
$\sigma=0.5$ & $0.90 \pm 0.03$ & $1.00 \pm 0.01$ & $0.86 \pm 0.03$ & $1.25 \pm 0.00$ & $1.25 \pm 0.00$ & 3 & 4 \\
$\sigma=1$   & $0.83 \pm 0.04$ & $1.00 \pm 0.00$ & $0.78 \pm 0.03$ & $5.00 \pm 0.01$ & $4.97 \pm 0.01$ & 3 & 5 \\
$\sigma=2$   & $0.87 \pm 0.04$ & $1.00 \pm 0.00$ & $0.83 \pm 0.04$ & $20.01 \pm 0.03$ & $19.89 \pm 0.03$ & 3 & 3 \\
\bottomrule
\end{tabular}
\label{tab:clust_selected_metrics}
\end{table}

\begin{table}[H]
\centering
\small
\caption{Three Clusters: Skeleton and DAG recovery metrics for different methods under varying disturbance levels ($\sigma$). (Small‐sample/Long‐series, $n=50,\;m=300$; mean $\pm$ 95\% CI over 50 runs, threshold $=0.02$).}
\label{tab:combined_recovery_metrics_three_clusters}
\setlength{\tabcolsep}{3pt}
\begin{tabular}{l|c|r|r|r|r|r}
\toprule
Method & Skel TPR & Skel FDR & Skel TNR & DAG TPR & DAG FDR & DAG TNR \\
\midrule
\multicolumn{7}{l}{\textbf{Disturbance Scale $\sigma = 0.5$}} \\
\midrule
Population       & $0.91 \pm 0.03$ & $0.47 \pm 0.02$ & $0.17 \pm 0.04$ & $0.62 \pm 0.04$ & $0.64 \pm 0.02$ & $0.63 \pm 0.02$ \\
Individual      & $1.00 \pm 0.00$ & $0.36 \pm 0.00$ & $0.40 \pm 0.01$ & $0.79 \pm 0.02$ & $0.49 \pm 0.01$ & $0.73 \pm 0.01$ \\
DAG-DC-ADMM              & $1.00 \pm 0.00$ & $0.19 \pm 0.03$ & $0.73 \pm 0.05$ & $0.92 \pm 0.02$ & $0.33 \pm 0.03$ & $0.82 \pm 0.02$ \\
Oracle   & $1.00 \pm 0.00$ & $0.08 \pm 0.02$ & $0.88 \pm 0.03$ & $0.88 \pm 0.03$ & $0.18 \pm 0.04$ & $0.92 \pm 0.02$ \\
\midrule
\multicolumn{7}{l}{\textbf{Disturbance Scale $\sigma = 1.0$}} \\
\midrule
Population       & $0.90 \pm 0.03$ & $0.48 \pm 0.02$ & $0.16 \pm 0.04$ & $0.61 \pm 0.04$ & $0.65 \pm 0.02$ & $0.62 \pm 0.02$ \\
Individual       & $1.00 \pm 0.00$ & $0.36 \pm 0.00$ & $0.40 \pm 0.01$ & $0.79 \pm 0.02$ & $0.49 \pm 0.01$ & $0.73 \pm 0.01$ \\
DAG-DC-ADMM              & $1.00 \pm 0.00$ & $0.15 \pm 0.02$ & $0.78 \pm 0.03$ & $0.93 \pm 0.01$ & $0.31 \pm 0.03$ & $0.84 \pm 0.02$ \\
Oracle   & $1.00 \pm 0.00$ & $0.09 \pm 0.02$ & $0.88 \pm 0.03$ & $0.87 \pm 0.03$ & $0.19 \pm 0.04$ & $0.92 \pm 0.02$ \\
\midrule
\multicolumn{7}{l}{\textbf{Disturbance Scale $\sigma = 2.0$}} \\
\midrule
Population       & $0.90 \pm 0.03$ & $0.47 \pm 0.02$ & $0.17 \pm 0.05$ & $0.60 \pm 0.04$ & $0.65 \pm 0.02$ & $0.62 \pm 0.02$ \\
Individual       & $1.00 \pm 0.00$ & $0.28 \pm 0.00$ & $0.57 \pm 0.01$ & $0.80 \pm 0.01$ & $0.42 \pm 0.01$ & $0.79 \pm 0.01$ \\
DAG-DC-ADMM              & $1.00 \pm 0.00$ & $0.09 \pm 0.02$ & $0.87 \pm 0.03$ & $0.96 \pm 0.01$ & $0.26 \pm 0.03$ & $0.87 \pm 0.02$ \\
Oracle   & $1.00 \pm 0.00$ & $0.08 \pm 0.02$ & $0.88 \pm 0.03$ & $0.88 \pm 0.03$ & $0.18 \pm 0.04$ & $0.92 \pm 0.02$ \\
\bottomrule
\end{tabular}
\end{table}

\begin{table}[!htbp]
\centering
\small
\caption{Three Clusters: Clustering Results (Large-sample/Short-series, $n=200$, $m=50$). Values are mean $\pm$ 95\% CI over 50 runs; 
The true number of clusters $R_{\text{truth}}=3$ is fixed, and estimated $\widehat{R}$ is reported as the mode over repetitions.}
\setlength{\tabcolsep}{3pt}
\begin{tabular}{lccccccc}
\toprule
Setting & ARI & Homogeneity & Completeness & $\text{Recon.\ Error}_{\text{truth}}$& $\text{Recon.\ Error}_{\text{est}}$ & $R_{\text{truth}}$ & $\widehat{R}$\\
\midrule
$\sigma=0.5$ & $0.36 \pm 0.04$ & $0.93 \pm 0.04$ & $0.40 \pm 0.02$ & $1.25 \pm 0.00$ & $1.25 \pm 0.00$ & 3 & 35 \\
$\sigma=1$   & $0.52 \pm 0.09$ & $0.98 \pm 0.01$ & $0.56 \pm 0.07$ & $5.00 \pm 0.01$ & $4.98 \pm 0.01$ & 3 & 5 \\
$\sigma=2$   & $0.74 \pm 0.08$ & $0.91 \pm 0.04$ & $0.72 \pm 0.05$ & $19.96 \pm 0.03$ & $19.95 \pm 0.08$ & 3 & 9 \\
\bottomrule
\end{tabular}
\label{tab:clust_selected_metrics_large}
\end{table}

\begin{table}[H]
\centering
\small
\caption{Three Clusters: Skeleton and DAG recovery metrics for different methods under varying disturbance levels ($\sigma$). (Large-sample/Small-series, $n=200,\;m=50$; mean $\pm$ 95\% CI over 50 runs, threshold $=0.02$).}
\label{tab:combined_recovery_metrics_three_clusters_large_sample_small_series}
\setlength{\tabcolsep}{3pt}
\begin{tabular}{l|c|r|r|r|r|r}
\toprule
Method & Skel TPR & Skel FDR & Skel TNR & DAG TPR & DAG FDR & DAG TNR \\
\midrule
\multicolumn{7}{l}{\textbf{Disturbance Scale $\sigma = 0.5$}} \\
\midrule
Population       & $0.90 \pm 0.04$ & $0.48 \pm 0.02$ & $0.16 \pm 0.05$ & $0.60 \pm 0.04$ & $0.65 \pm 0.02$ & $0.62 \pm 0.02$ \\
Individual       & $0.94 \pm 0.00$ & $0.39 \pm 0.00$ & $0.36 \pm 0.00$ & $0.65 \pm 0.01$ & $0.58 \pm 0.01$ & $0.69 \pm 0.00$ \\
DAG-DC-ADMM                & $1.00 \pm 0.00$ & $0.23 \pm 0.03$ & $0.65 \pm 0.05$ & $0.87 \pm 0.02$ & $0.40 \pm 0.03$ & $0.77 \pm 0.02$ \\
Oracle   & $1.00 \pm 0.00$ & $0.13 \pm 0.02$ & $0.81 \pm 0.03$ & $0.87 \pm 0.03$ & $0.24 \pm 0.03$ & $0.89 \pm 0.02$ \\
\midrule
\multicolumn{7}{l}{\textbf{Disturbance Scale $\sigma = 1.0$}} \\
\midrule
Population       & $0.89 \pm 0.04$ & $0.48 \pm 0.02$ & $0.15 \pm 0.04$ & $0.60 \pm 0.04$ & $0.66 \pm 0.02$ & $0.62 \pm 0.02$ \\
Individual       & $0.88 \pm 0.01$ & $0.28 \pm 0.00$ & $0.63 \pm 0.00$ & $0.60 \pm 0.01$ & $0.51 \pm 0.01$ & $0.79 \pm 0.00$ \\
DAG-DC-ADMM                & $0.99 \pm 0.00$ & $0.17 \pm 0.02$ & $0.75 \pm 0.03$ & $0.88 \pm 0.02$ & $0.36 \pm 0.03$ & $0.81 \pm 0.02$ \\
Oracle   & $1.00 \pm 0.00$ & $0.13 \pm 0.02$ & $0.81 \pm 0.03$ & $0.86 \pm 0.03$ & $0.25 \pm 0.04$ & $0.89 \pm 0.02$ \\
\midrule
\multicolumn{7}{l}{\textbf{Disturbance Scale $\sigma = 2.0$}} \\
\midrule
Population       & $0.91 \pm 0.03$ & $0.48 \pm 0.02$ & $0.16 \pm 0.04$ & $0.59 \pm 0.03$ & $0.66 \pm 0.02$ & $0.62 \pm 0.02$ \\
Individual       & $0.96 \pm 0.00$ & $0.43 \pm 0.00$ & $0.27 \pm 0.00$ & $0.66 \pm 0.01$ & $0.60 \pm 0.01$ & $0.66 \pm 0.00$ \\
DAG-DC-ADMM                & $0.99 \pm 0.01$ & $0.16 \pm 0.04$ & $0.75 \pm 0.06$ & $0.86 \pm 0.02$ & $0.27 \pm 0.05$ & $0.86 \pm 0.03$ \\
Oracle   & $1.00 \pm 0.00$ & $0.12 \pm 0.02$ & $0.83 \pm 0.04$ & $0.88 \pm 0.03$ & $0.22 \pm 0.04$ & $0.90 \pm 0.02$ \\
\bottomrule
\end{tabular}
\end{table}